\documentclass[acmlarge]{acmart}

\AtBeginDocument{%
	\providecommand\BibTeX{{%
			\normalfont B\kern-0.5em{\scshape i\kern-0.25em b}\kern-0.8em\TeX}}}

\usepackage{xspace}
\usepackage{comment}
\usepackage{multirow}
\usepackage{amsmath}
\usepackage{amsfonts}
\usepackage{makecell}
\usepackage{xcolor}
\usepackage{footnote}
\makesavenoteenv{tabular}
\makesavenoteenv{table}

\newcommand{\etal}{et al.\@\xspace}

\begin{document}
	
\title[Neural Snapshot Ensemble]{NeuSE: A Neural Snapshot Ensemble Method for Collaborative Filtering}
	

	\author{Dongsheng Li}
	\affiliation{%
		\institution{Fudan University}
		\streetaddress{2005 Songhu Road}
		\city{Yangpu District}
		\state{Shanghai}
		\country{China}}
	\email{dongshengli@fudan.edu.cn}
	\authornote{Dongsheng Li and Haodong Liu contributed equally to this work.}

	\author{Haodong Liu}
	\affiliation{%
		\institution{University of Electronic Science and Technology of China}
		\streetaddress{2006 Xiyuan Ave}
		\city{Chengdu}
		\state{Sichuan}
		\country{China}}
	\email{haodong.lhd@gmail.com}
	\authornotemark[1]
	
	\author{Chao Chen}
	\affiliation{%
		\institution{Shanghai Jiao Tong University}
		\streetaddress{800 Dongchuan Road}
		\city{Minhang District}
		\state{Shanghai}
		\country{China}}
	\email{chao.chen@sjtu.edu.cn}
	
	\author{Yingying Zhao}
	\affiliation{\institution{Fudan University}
		\streetaddress{2005 Songhu Road}
		\city{Yangpu District}
		\state{Shanghai}
		\country{China}}
	\email{yingyingzhao@fudan.edu.cn}
	\authornote{Yingying Zhao and Bo Yang are the corresponding authors.}
	
	\author{Stephen M. Chu}
	\affiliation{%
		\institution{IBM Research - China}
		\streetaddress{55 Chuanhe Road}
		\city{Pudong New District}
		\state{Shanghai}
		\country{China}}
	\email{schu@cn.ibm.com}
	
	\author{Bo Yang}
	\affiliation{%
		\institution{University of Electronic Science and Technology of China}
		\streetaddress{2006 Xiyuan Ave}
		\city{Chengdu}
		\state{Sichuan}
		\country{China}}
	\email{yangbo@uestc.edu.cn}
	\authornotemark[2]

\renewcommand{\shortauthors}{Li, et al.}

\begin{abstract}
In collaborative filtering (CF) algorithms, the optimal models are usually learned by globally minimizing the empirical risks averaged over all the observed data. However, the global models are often obtained via a performance tradeoff among users/items, \textit{i.e.}, not all  users/items are perfectly fitted by the global models due to the hard non-convex optimization problems in CF algorithms. Ensemble learning can address this issue by learning multiple diverse models but usually suffer from efficiency issue on large datasets or complex algorithms. In this paper, we keep the intermediate models obtained during global model learning as the snapshot models, and then adaptively combine the snapshot models for individual user-item pairs using a memory network-based method. Empirical studies on three real-world datasets show that the proposed method can extensively and significantly improve the accuracy (up to 15.9\% relatively) when applied to a variety of existing collaborative filtering methods.
\end{abstract}


\maketitle

\section{Introduction}

Collaborative filtering (CF) is the most important class of methods in today's recommender systems~\cite{survey05,aggarwal2016recommender}. In many existing CF algorithms, \textit{e.g.}, matrix factorization (MF), the models are usually learned via minimizing empirical risks. Since the loss functions usually aggregate  the losses over all users and items, convergences are reached when most users and items are appropriately modeled but not all of them~\cite{Beutel17,li2017mrma}. Our case studies show that the global models can perfectly fit some users/items, but the other users/items may suffer from either overfitting or underfitting. Recently, Beutel~\etal~\shortcite{Beutel17} also proved that globally optimized models are typically not the best models for every subset of the data. In summary, the globally optimized models often yield suboptimal performances in CF  problems~\cite{Beutel17,lee2013local,li2017mrma}.

One may imagine that adaptive learning rate methods, \textit{e.g.}, AdaGrad~\cite{adagrad}, can address this issue by manipulating the learning of individual users/items to achieve a ``truly'' global optimum. However, the optimization problem in CF algorithms are generally non-convex due to the co-existence of users and items~\cite{rennie2005fast,li2017ermma}, so that adaptive learning rate methods can only help to better converge to one local optimum but cannot explore other local optimums that may achieve better performance as demonstrated in our case studies. Another kind of possible solutions may be Bayesian approaches~\cite{BPMF,FM}, which perform approximate inference by integrals from sampling. However, Bayesian approaches generally face two main challenges~\cite{BPMF}: 1) choosing the prior distributions for the unknown model parameters is non-trivial and 2) these methods are usually computationally demanding and thus limited to small-scale problems. Therefore, ensemble learning-based CF algorithms~\cite{lee2013local,li2017mrma}, which learn multiple base models each of which can achieve better performance over a subset of data, are more appropriate to address the suboptimal performance issue of global models~\cite{Beutel17}. However, the learning of multiple base models in ensemble methods will suffer from efficiency issue on large datasets, \textit{e.g.}, the Yahoo! Music dataset~\cite{Yahoo}, or on very complex models, \textit{e.g.}, deep neural networks~\cite{NCF}. Therefore, it is desirable to design ensemble methods to address the suboptimal performance issue of global models without significantly increasing the computational overhead.

Following recent works that create ensembles from the slices of the learning trajectory~\cite{committee,snapshot}, we also combine the snapshot models obtained during global model learning, so that ensemble learning can be achieved with little additional training costs. Different from existing works which simply average the snapshot models~\cite{committee,horizontal,snapshot}, we propose a {\bf Neu}ral network-based {\bf S}napshot {\bf E}nsemble (NeuSE) method, in which three memory networks are introduced to weigh different snapshot models: 1) model memory network to capture the global importance of each snapshot model,  2) user memory network to capture the user-wise importance of each snapshot model and 3) item memory network to capture the item-wise importance of each snapshot model. Then, a fully connected layer with three mapping modules are adopted to combine the outputs of memory networks to produce adaptive ensemble weights for different user-item pairs. To achieve better generalization performance, we propose a soft label-based  learning method, which can reduce the loss when the predicted snapshot model is close to the optimal snapshot model and increase the loss otherwise. Empirical studies on several real-world datasets demonstrate that the proposed method can extensively improve the performance of existing collaborative filtering methods (up to 15.9\% relatively) on almost all kinds of algorithms and recommendation tasks.

\section{Related Work}

Collaborative filtering (CF) has been extensively studied in recent years and the state-of-the-art performances are achieved via  machine learning techniques, \textit{e.g.}, supervised learning~\cite{FM,Zhang2019} and unsupervised learning~\cite{koren2009matrix,WANG2017102,KARABADJI2018153}. However, due to the non-convex nature of CF problems, many existing algorithms cannot learn the ``optimal'' model~\cite{lee2013local,Beutel17,li2017mrma}, \textit{i.e.}, the globally optimized models are typically not optimal for each subset of the data~\cite{Beutel17}. Therefore, ensemble learning~\cite{lee2013local,li2017mrma}, which learn multiple base models each of which can achieve better performance over a subset of data, are more appropriate to address the suboptimal performance issue~\cite{Beutel17}. However, efficiency issue may arise in ensemble methods due to the learning of multiple models, especially on large datasets~\cite{snapshot} or complex neural network methods~\cite{NCF}. In addition, many existing ensemble collaborative filtering methods were proposed for specific CF methods and thus may not be applied to other CF methods, \textit{e.g.}, LLORMA~\cite{lee2013local} achieved excellent ensemble performance on matrix factorization methods but cannot be applied to UserKNN or ItemKNN methods. Different from existing ensemble CF methods, \textit{the goal of this work is to improve the performances of almost all kinds of existing CF algorithms without significantly increasing the training overheads}.

Many techniques have been proposed to learn better machine learning models in general tasks, \textit{e.g.}, adaptive learning rate~\cite{adagrad,adaerror}, Dropout~\cite{Dropout}, and Bayesian approaches~\cite{FM,BPMF,xiao2019bayesian}, \textit{etc}. However, adaptive learning rate methods are not capable of exploring different local optimums to achieve better performance. Dropout can indeed improve the performances of neural network-based CF methods~\cite{DropoutNet} but cannot be applied to non-neural network methods, \textit{e.g.}, matrix factorization~\cite{koren2009matrix}.  Factorization models learned via Bayesian treatment~\cite{FM,BPMF}, \textit{e.g.}, MCMC, achieved superior performance over those learned via SGD or ALS. But as mentioned earlier, Bayesian approaches usually suffer from unknown prior distribution and scalability issues. In addition, \textit{the proposed method is orthogonal to the above methods and thus can be combined with them to further improve performance as demonstrated in the experiments.}

This work follows existing works that create ensembles from the slices of learning trajectory. Swann~\etal~\shortcite{committee} and Xie~\etal~\shortcite{horizontal} both suggested to average the outputs of snapshot models during stable range of epochs to improve model performance. Huang~\etal~\shortcite{snapshot} proposed to learn multiple neural network models converging to different local minimums during one optimization path via cyclic learning rate schedules. However, simply averaging the outputs of all snapshot models may raise challenges to snapshot model selection. As pointed out by Swann~\etal~\shortcite{committee}, models with poor generalization ability should not be selected, because averaging poor models may affect the ensemble performance. Therefore, the ensemble method should be capable of weighing the importance of different snapshot models so that poor models can be excluded most of the time. Xie~\etal~\shortcite{horizontal} also adopted a simple MLP on top of the snapshot models to learn the ensemble weights of different classifiers. However, their method only captures the global importance of snapshot models, which is not enough in collaborative filtering tasks where each user may have varying  ensemble weights across different items. Our empirical studies show that a simple MLP can only achieve similar performance with simple averaging the outputs of the snapshot models. Different from the above works, \textit{the proposed method can adaptively change the ensemble weights for different user-item pairs, so that a user may have different ensemble weights on different items to better characterize the non-convexity of collaborative filtering.}

\section{Problem Formulation}
This section first analyzes the non-convex optimization problem in
collaborative filtering. Then, we motivate the snapshot ensemble
problems by three case studies.

\subsection{Preliminaries}
For many machine learning-based CF methods, due to the coexistence of users and items, both user parameters and item parameters should be learned via minimizing predefined loss functions over the entire dataset. Then, an interaction function between user parameters and item parameters can be adopted to generate recommendation scores, \textit{e.g.}, matrix factorization methods use dot-product as the interaction function~\cite{koren2009matrix} and NeuMF uses multi-layer perceptron as the interaction function~\cite{NCF}. Given the observed rating matrix $R$ and a loss function $\mathcal{L}$, the optimal user parameters ($U^*$) and item parameters ($V^*$) are obtained as follows:
\begin{equation}
\label{eqn:loss}
\textstyle
U^*,V^* = \arg\min_{U,V}\mathcal{L}(R|U,V). 
\end{equation}
To learn the globally optimal $U,V$, iterative learning methods, \textit{e.g.}, SGD~\cite{koren2009matrix}, ALS~\cite{Hu08,FM} and MCMC~\cite{FM}, are usually adopted. $\mathcal{L}$ can be convex to $U$ or $V$ alone, \textit{e.g.}, in the mean square loss function, but it is usually non-convex to $U,V$ together. Therefore, many local optimums exist due to the non-convex optimization problem. Although it has been proved that any local optimum is also global optimum in matrix factorization problem under certain assumption~\cite{Ge17}, \textit{i.e.}, on symmetric positive definite matrices, there are no such results for general CF problems to the best of our knowledge because user-item rating matrices in CF tasks are asymmetric.

\begin{figure*}[tb!]
	\begin{minipage}[t]{0.66\linewidth}
		\centering
		\includegraphics[width=0.495\textwidth]{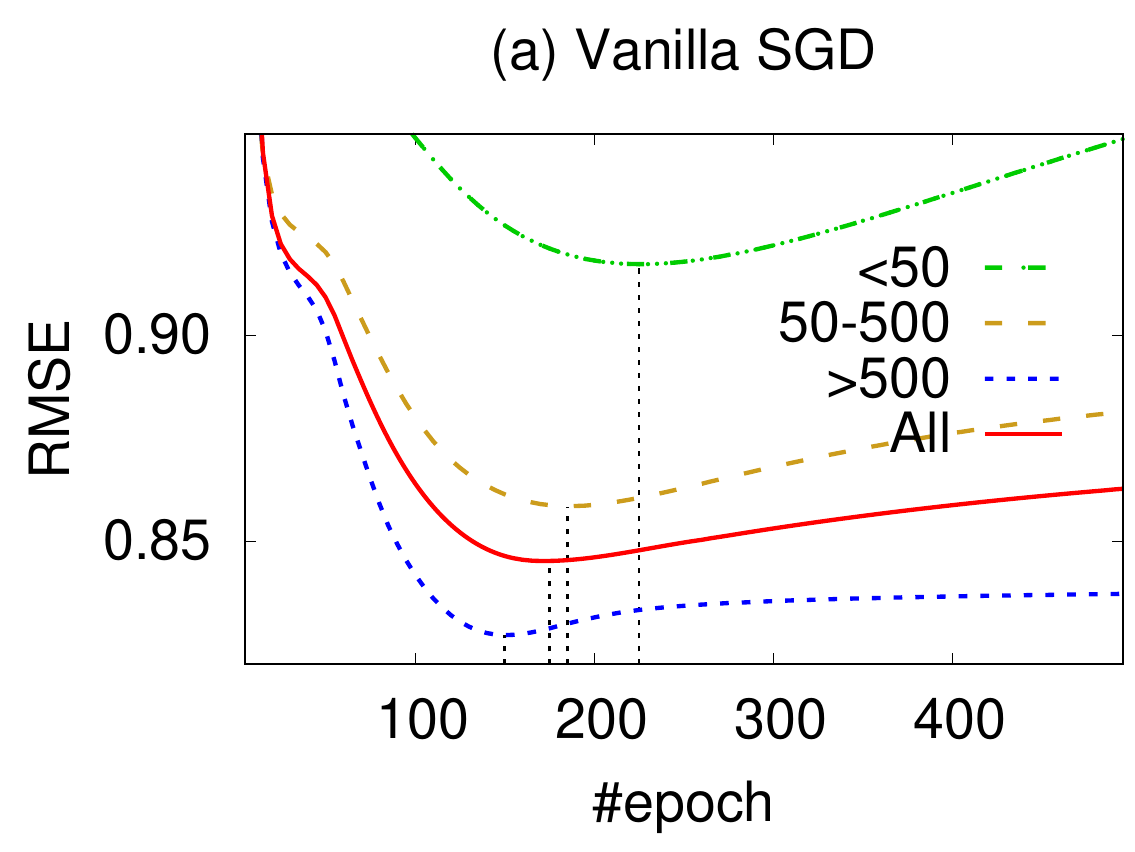}
		\includegraphics[width=0.495\textwidth]{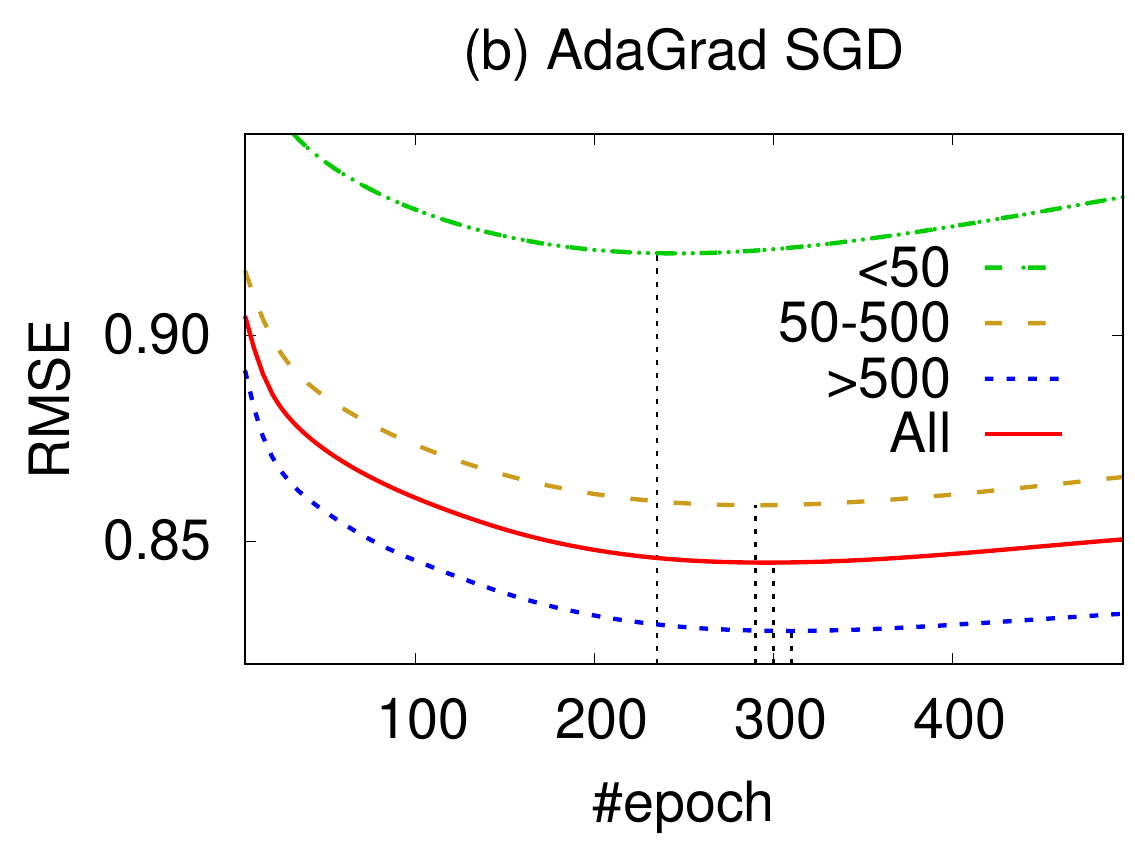}
		\caption{Test RMSE trends on rating prediction task for users with different number of ratings using the RSVD method~\cite{paterek2007improving}, \textit{i.e.}, ``$<$50'' means users with less than 50 ratings and so forth. We train the model using vanilla SGD in figure (a) and using SGD with adagrad~\cite{adagrad} in figure (b).
			\label{fig:test_error_rating}}
	\end{minipage}
	\hfill
	\begin{minipage}[t]{0.33\linewidth}
		\centering
		\includegraphics[width=0.99\textwidth]{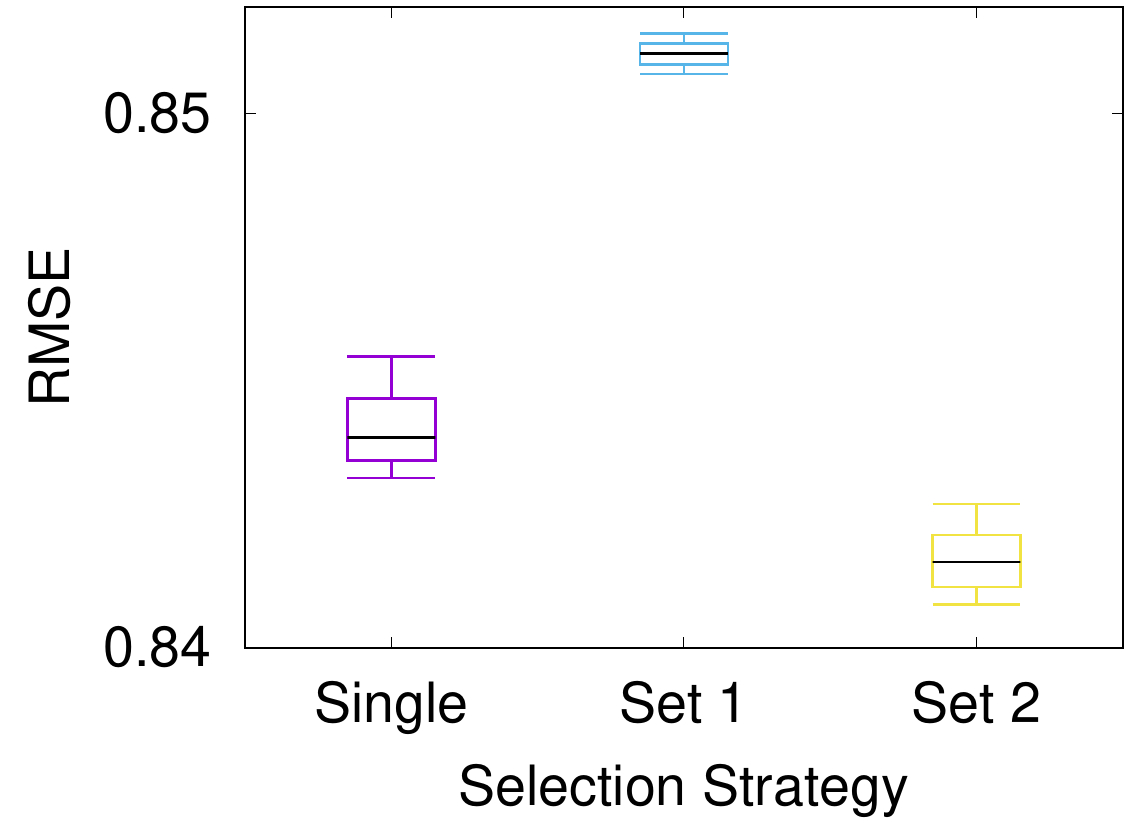}
		\caption{Performance of three strategies:
			1) single (the optimal snapshot model),
			2) set 1 (epoch = 50, 100, 150,..., 500),
			3) set 2 (epoch = 100, 200, 300).
			\label{fig:accuracy_selection}}
	\end{minipage}
\end{figure*}

\subsection{Case Studies}

This section presents three case studies on the MovieLens 1M dataset to motivate the proposed method, in which we randomly split the data into training and test sets by 9:1. In Figure~\ref{fig:test_error_rating} and Figure~\ref{fig:accuracy_selection}, we use one of the classic matrix factorization methods -- the RSVD method~\cite{paterek2007improving} in the rating prediction task.

\subsubsection{Global vs. Local Convergence}

Figure~\ref{fig:test_error_rating} shows the test errors of different kinds of users with increasing number of epochs in the rating prediction task. As shown in Figure~\ref{fig:test_error_rating} (a), users with different numbers of ratings converge at different epochs and the global optimum is achieved with epoch around 175 which is near the optimum for the majority of users (with 50-500 ratings). Therefore, users with less than 50 ratings and users with more than 500 ratings will suffer from underfitting and overfitting, respectively, if we only use the globally optimal model. Similar phenomenon can also be observed from Figure~\ref{fig:test_error_rating} (b), in which the model is trained with AdaGrad SGD instead of vanilla SGD in Figure~\ref{fig:test_error_rating} (a). Due to the co-existence of users and items, the optimal snapshot models are also varying for different kinds of items. Therefore, the optimal snapshot model for a user could vary across different items (account for about 90\% of ratings in the case study). This case study confirms that a globally optimal model is typically not the optimum for all subsets of the data. Therefore, it is necessary to choose different snapshot models for different user-item pairs to achieve better accuracy.

It should be noted that the test errors for different kinds of users in Figure~\ref{fig:test_error_rating} (b) are relatively more stable than those in Figure~\ref{fig:test_error_rating} (a), which seems that the suboptimal performance issue of collaborative filtering can be alleviated using adaptive learning rate methods. However, we can also observe in this study that the optimal (global) model trained via AdaGrad SGD achieved almost the same test RMSE compared with the optimal (global) model trained via vanilla SGD. The same phenomenon has also be confirmed by several recent works~\cite{wilson2017marginal,smith2017don}, in which marginal improvements are observed on adaptive learning rate methods in several machine learning tasks. Therefore, we claim that adaptive learning rate methods can help to easier determine the optimal (global) model, i.e., near-optimal models can be obtained in a relatively large range of epochs, but these methods cannot guarantee superior performance compared to conventional learning methods and thus cannot solve the suboptimal performance issue fundamentally.

\subsubsection{Model Selection}

In ensemble learning, the performance may degrade if poor base models are selected~\cite{Zhou02}. Therefore,  snapshot model selection is crucial to existing methods~\cite{committee}. Xie~\etal~\shortcite{horizontal} suggested to select snapshots from ``a stable range of epoch'', which is not applicable to some CF problems where stable ranges are unavailable (\textit{e.g.}, Figure~\ref{fig:test_error_rating} (a)). Figure~\ref{fig:accuracy_selection} compares the accuracy of three selection strategies: 1) single (the globally optimal snapshot model), 2) set 1 (all snapshot models from every 50 epochs till epoch = 500) and 3) set 2 (snapshot models with epoch = 100, 200, 300). Here, we report the test RMSE of a simple average over all snapshot models with ten random data splits. As shown in the figure, if the snapshot models are not properly selected (set 1), \textit{i.e.}, too noisy, the ensemble performance of averaging the snapshot models is even worse than the optimal single model. On the contrary, if the snapshot models are properly selected (set 2), a simple average can achieve observable accuracy improvement, \textit{e.g.}, around 0.3\% relatively in this study. This study indicates that an ensemble learner is needed to assign higher weights to the superior models and lower weights for inferior models for different user-item pairs to overcome the challenge of snapshot model selection.

\section{NeuSE: Neural Snapshot Ensemble}



The proposed method consists of two key components:
\begin{itemize}
\item {\bf Snapshot Selection}. Due to the non-convex optimization problems in CF, it is challenging to intentionally learn several locally optimal models. As shown in the case study, each intermediate model in the learning trajectory could be optimal for a subset of users and items, which should be all considered in the ensemble. Thus, we propose to select snapshot models with fixed epoch variations ($\Delta T$), \textit{i.e.}, we keep the snapshot models for every $\Delta T$ epochs during training. For non-machine learning-based CF methods, we can also select intermediate models as snapshots, \textit{e.g.}, models with different number of neighbors in memory-based methods.

\item {\bf Ensemble Learning}. For each user-item pair, each snapshot model will have a predicted rating, all of which are used as the input of the snapshot ensemble method.  The ensemble method adopts a multi-memory network to capture the global, user-specific and item-specific importance of the snapshot models via different memory networks. To ensure robust ensemble learning, we change the hard label (optimal snapshot model as 1 and 0 for all others) to soft labels, because nearby snapshot models of the optimal model could have much smaller error than faraway snapshot models (as shown in Figure~\ref{fig:test_error_rating}). After learning the ensemble model, we can use its output as the adaptive ensemble weights.
\end{itemize}

\subsection{Network Architecture}
\begin{figure}
	\includegraphics[width=0.66\textwidth]{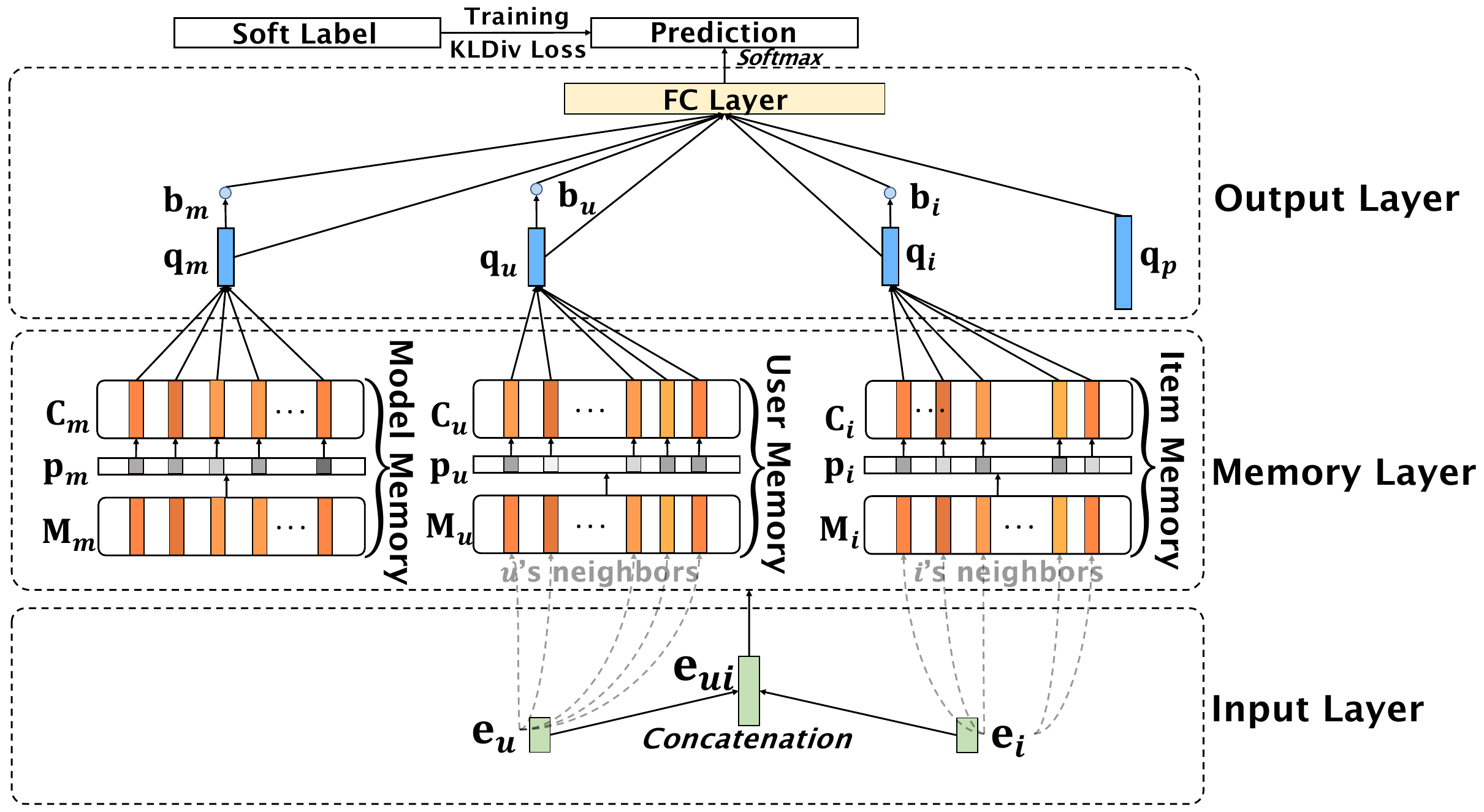}
	\caption{Architecture of the proposed multi-memory network for snapshot ensemble with one memory layer.}
	\label{fig:arch}
\end{figure}

The key step in NeuSE is to adaptively generate the ensemble weights for different ratings. However, the ensemble model could easily overfit if we only consider the limited ratings of individual users/items. To remedy this, we adopt the memory networks~\cite{weston2014memory} in NeuSE, which can maintain long-term memories to facilitate inference. The memory components store the weight vectors of global information and the top $k$ neighbors of each user/item, which can help to address the limited data issue of individual users and items by leveraging the global information and user/item neighborhood information.
As illustrated in Figure~\ref{fig:arch}, the proposed ensemble network consists of three layers: 1) input layer, 2) memory layer, and 3) output layer.

\subsubsection{Input Layer}
We use the identities of a user and an item as the input feature via one-hot encoding, \textit{i.e.}, we represent the $u$-th user with a sparse vector in which only the $u$-th element is 1 and all other elements are 0. The sparse representation of a user or an item is then projected to a dense vector $\mathbf{e}_u  \in  \mathbb{R}^d$ or $\mathbf{e}_i  \in  \mathbb{R}^d$ using one fully connected layer, which can be seen as the latent vector for the user or item. Given $D=2d$, we concatenate the two embedding vectors to construct $\mathbf{e}_{ui}  \in  \mathbb{R}^D$, which is then fed into the subsequent memory layer.

\subsubsection{Memory Layer}
The memory layer consists of three components: model memory, user memory and item memory.

\noindent\textit{Model Memory}.
The model memory consists of an internal memory matrix $\mathbf {M} _ { m }  \in  \mathbb { R } ^ { D \times N_m }$ and an external memory matrix $\mathbf{C} _ m  \in  \mathbb {R}^{D \times N_m}$, where $N_m$ represents the number of snapshot models and $D$ denotes the dimensionality of each memory cell. Each snapshot model $s$ is embedded in two memory slots $\mathbf{m}_{s}  \in  \mathbf{M}_m$ and $\mathbf{c}_{s}  \in  \mathbf{C}_m$, storing the specific internal and external attributes of it, respectively. For each snapshot model $s$, the corresponding internal memory slot $\mathbf{m}_{s}$ is used to measure the relevance between $s$ and user $u$ and item $i$, \textit{i.e.}, if $\mathbf{m}_{s}$ is similar to $\mathbf{e}_{ui}$ then we know that model $s$ is important to predict $u$'s rating on $i$. In other words, $\mathbf{m}_{s}$ captures how many user-item pairs can use model $s$ to accurately predict their ratings. Different from internal memory, each slot in the external memory, \textit{e.g.}, $\mathbf{c}_{s}$, can be regarded as the representation of each snapshot model.

More specifically, we form a snapshot model preference vector $\mathbf{p}_s^{M}  \in  \mathbb{R}^D$ -- each dimension $p_{uis}^M$ is the level of compatibility among snapshot model $s$, user $u$ and item $i$ as follows:
\begin{equation}
p _ { u i s }^M=  \mathbf{m} _ { s } ^ \mathrm{ T } \mathbf{e} _ { u i }, \quad \forall s \in N(m), 
\end{equation}
where $N(m)$ represents the set of snapshot models. In other words, $p_{uis}^M$ measures the importance of snapshot model $s$ for the rating made by user $u$ on item $i$.

To adaptively weigh snapshot models, we need to predefine some heuristic functions, \textit{e.g.}, cosine similarity, to distinguish superior snapshot models from inferior ones. Therefore, we introduce the attention mechanism, in which we can learn an adaptive weighting function to focus on only a subset of the influential snapshot models to derive the final score. More formally, we compute the attention weight for a given snapshot model to infer its unique contribution as follows:
\begin{equation}
w _ { u i s }^M =  \frac{ \exp \left( p _ { u i s } ^M\right) }{ \sum _ { k \in N(m) } \exp \left( p _ { u i k }^M \right) }, \quad \forall s \in N(m). 
\end{equation}
The attention mechanism allows our model to place higher weights on appropriate snapshot models and lower weights on ones that may be less compatible with user $u$ and item $i$. Then, we can construct the final representation of model memory by combining the external model memory with the attention weights as follows:
\begin{equation}
\label{eqn:model_mem}
\textstyle
\mathbf { q } _ { m } = \sum _ { s \in N(m) } w _ { u i s }^M \mathbf { c } _ { s }.
\end{equation}
Here, $\mathbf{c}_s$ is the embedding vector for snapshot model $s$ in the external model memory, which stores the long-term information pertaining specifically to each snapshot model. In Equation~\ref{eqn:model_mem}, the attention mechanism can selectively weigh the snapshot models according to the targeted user and item, and the output of model memory $\mathbf{q}_m$ represents a weighted sum over the snapshot models composed of the compatibility among the targeted user, item and all snapshot models.

\noindent\textit{User Memory and Item Memory}.
Similar to model memory, the user memory and item memory both consist of an internal memory and an external memory to store the  long-term memories of the snapshot models for each user and item, respectively. The memory components work as neighborhood aggregation function to select similar users/items via neural attention mechanism to mitigate the insufficient data issue of individual users/items. For each user $v$ in the neighborhood, we form user $u$'s preference vector $\mathbf{p}^U_u  \in  \mathbb{R}^D$ in which each dimension $p_{uiv}^U$ measures the level of similarity between the target user $u$ and its neighbor $v$ on the given item $i$ as follows:
\begin{equation}
p _ { u i v }^U =  \mathbf{m} _ { v } ^ \mathrm{ T } \mathbf{e} _ { u i }, \quad \forall v \in N(i), 
\end{equation}
where the neighborhood $N(i)$ represents the set of all users who have interactions with item $i$, and $\mathbf{m}_v$ is the embedding vector of user $v$ in the internal user memory matrix $\mathbf{M}_u \in \mathbb {R}^{D \times N_i}$, where $N_i$ denotes the number of users in $N(i)$. Then, the weight of each neighbor $v$ on item $i$ can be described as follows:
\begin{equation}
w _ { u i v }^U = \frac { \exp \left( p _ { u i v }^U \right) } { \sum _ { k \in N(i) } \exp \left( p _ { u i k } ^U\right) }, \quad \forall v \in N(i). 
\end{equation}
$w_{ui*}^U$ produces a distribution over the neighborhood of $u$ on item $i$. Then, we can construct the final neighborhood-based representation of user $u$ as follows:
\begin{equation}
\textstyle
\mathbf { q } _ { u } = \sum _ { v \in N(i) } w _ { u i v }^U \mathbf { c } _ { v },  
\end{equation}
where $\mathbf{c}_v$ is the memory slot for user $v$ in the external user memory matrix $\mathbf{C}_u \in \mathbb{R}^{D \times N_i}$.

The item memory network is similarly designed as the user memory network. Given an item $j$ in the neighborhood $N(u)$ --- the set of all items having interactions with user $u$, we can describe the process in item memory as follows:
\begin{equation}
p _ { u i j }^I =  \mathbf{m} _ { j } ^ \mathrm{ T } \mathbf{e} _ { u i }, \quad \forall j \in N(u), 
\end{equation}
where $\mathbf{m}_j$ is the embedding vector of item $j$ in the internal item memory matrix $\mathbf{M}_i \in \mathbb {R}^{D \times N_u}$. $N_u$ denotes the number of items in $N(u)$. Then, the weight of each neighboring item $j$ can be described as follows:
\begin{equation}
w _ { u i j }^I = \frac { \exp \left( p _ { u i j }^I \right) } { \sum _ { k \in N(u) } \exp \left( p _ { u i k }^I \right) }, \quad \forall j \in N(u). 
\end{equation}
Finally, we construct the neighborhood-based representation of item $i$ as follows:
\begin{equation}
\textstyle
\mathbf { q } _ { i } = \sum _ { j \in N(u) } w _ { u i j }^I \mathbf { c } _ { j },  
\end{equation}
where $\mathbf{c}_j$ is the memory slot for item $j$ in the external item memory matrix $\mathbf{C}_i \in \mathbb{R}^{D \times N_u}$.

\subsubsection{Output Layer}
The output layer takes the predictions of all snapshot models and the outputs of the memory networks to make the final prediction. Similar to Biased MF~\cite{koren2009matrix} and FM~\cite{FM}, we introduce global bias, user bias and item bias by a linear projection from their latent representations ($\mathbf{q}_{m}, \mathbf{q}_{u}, \mathbf{q}_{i}$), which are helpful to capture the unique characteristics of data, user and item, respectively.

We project the final representations of the three memory networks to global/user/item biases as follows:
\begin{eqnarray}
b_m &=& \mathbf{W}_{model}^\mathrm{T} \mathbf{q}_m + \mathrm{b}_{model}, \\
b_u &=& \mathbf{W}_{user}^\mathrm{T} \mathbf{q}_u + \mathrm{b}_{user}, \\
b_i &=& \mathbf{W}_{item}^\mathrm{T} \mathbf{q}_i + \mathrm{b}_{item}. 
\end{eqnarray}
$\mathbf{W}_{model}, \mathbf{W}_{user}, \mathbf{W}_{item}  \in  \mathbb{R}^{D \times 1}$ and $\mathrm{b}_{model}, \mathrm{b}_{user}, \mathrm{b}_{item}$ are parameters to be learned.

Then, we concatenate the outputs of all snapshot models ($\mathbf{q}_p$), the latent representations from the three memory networks ($\mathbf{q}_m, \mathbf{q}_u, \mathbf{q}_i$) and the three bias terms above ($b_m, b_u, b_i$) together, and feed them into a fully connected layer with $\mathrm{softmax}$ activation function  as follows:
\begin{equation}
\hat {\mathbf{y}} = \mathrm{softmax}(\mathbf{W}^\mathrm{T}
\left[
\mathbf{q}_p, \mathbf{q}_m, \mathbf{q}_u, \mathbf{q}_i, b_m, b_u, b_i
\right]
+\mathbf{b}). 
\end{equation}
$\mathbf{W}  \in  \mathbb{R}^{(N_m + 3D + 3) \times N_m}$ and $\mathbf{b} \in \mathbb{R} ^ {N_m}$ are the parameters.

Finally, we obtain the prediction $\hat {\mathbf{y}}  \in  \mathbb{R}^{N_m}$, which represents the ensemble weights for the snapshot models considering not only the predictions from all snapshot models but also global attributes of all snapshot models and local information from the neighborhoods of users and items.

\begin{figure}
	\includegraphics[width=0.66\textwidth]{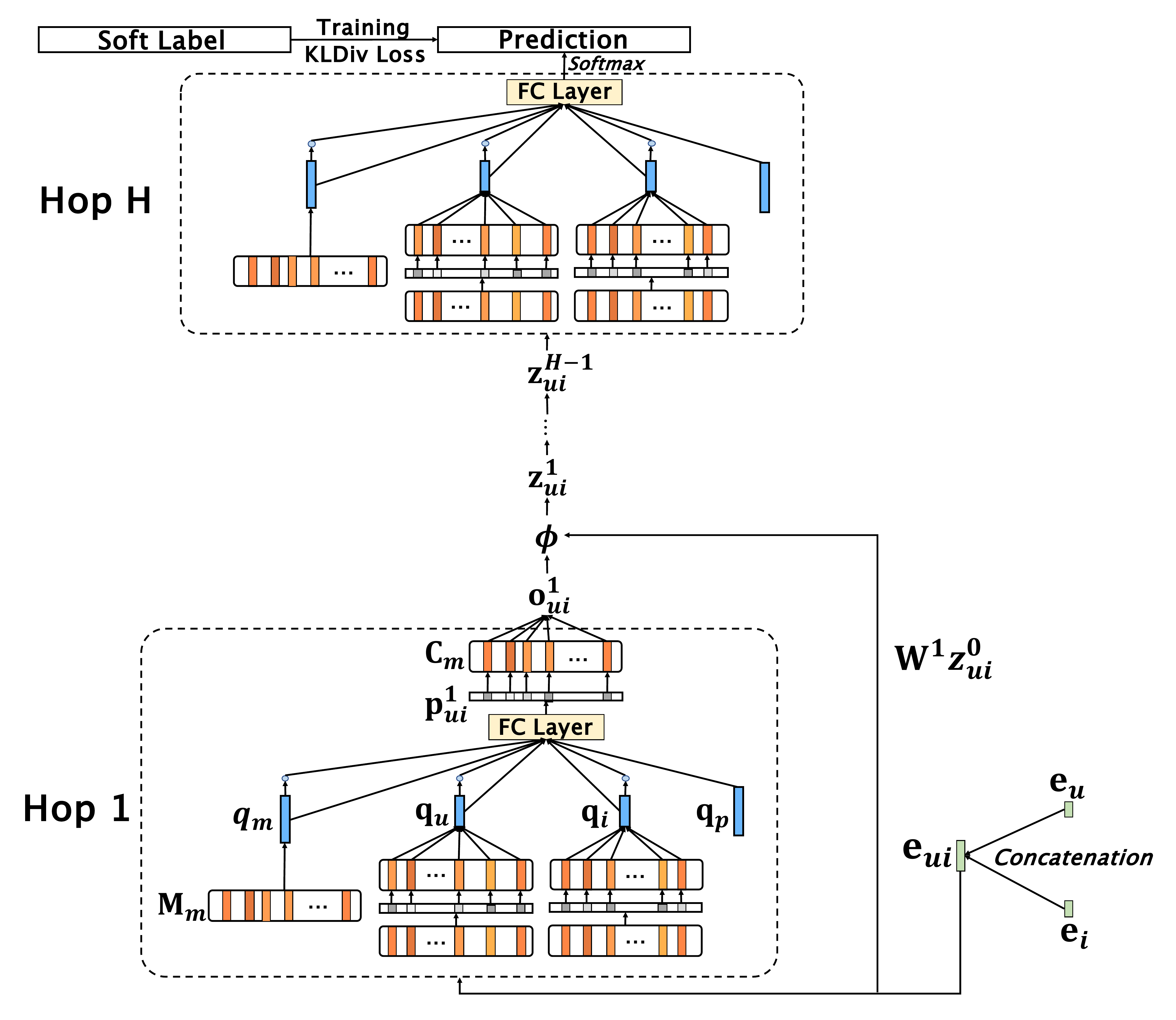}
	\caption{Architecture of the proposed deep multi-memory network for snapshot ensemble with multiple memory layers. Each hop denotes one layer of model/user/item memory networks with feature mapping layers.  The neighborhood information is the same as that in Figure~\ref{fig:arch}, which is omitted for simplicity.}
	\label{fig:arch_deep}
\end{figure}

\subsubsection{Multi-Layer Memory Networks}
We now illustrate how to extend the proposed method to incorporate multiple memory layers in Figure~\ref{fig:arch_deep}. Unlike the single hop memory network in Figure~\ref{fig:arch}, multi-layer memory modules can utilize the neighborhood representation of users and items in previous layers to derive the representation of snapshot models in the current layer and vice versa. In other words, the model has the chance to refine snapshot model representation by looking back and reconsidering the neighborhood information and vice versa. More specifically, multi-layer memory modules are stacked together in which the output of the $h^{th}$ hop is used as the input to the $(h+1) ^ {th}$ hop. Each hop $h$ (except for the last hop) queries the internal model memory, user memory and item memory via the attention mechanism and combines the predictions from all snapshot models to derive the attention weight vector $\mathbf{p} _ {ui} ^ h$, in which each dimension $p _ {uis} ^ h$ is the importance of snapshot model $s$ in hop $h$:
\begin{equation}
\mathbf{p} _ {ui} ^ h = \mathrm{softmax}
(\mathbf{W} _ h ^ \mathrm{T} \left[
\mathbf{q} _ m,
\mathbf{q} _ u,
\mathbf{q} _i ,
\mathbf{q} _ p,
b_m,
b_u,
b_i
\right]
+ \mathbf{b} _ h). 
\end{equation}
Here, $\mathbf{W} _ h$ is a square weight matrix mapping the concatenated vector to the latent space. Next, we can construct the output of hop $h$ as follows:
\begin{equation}
\textstyle
\mathbf{o} _ {ui} ^ h = \sum _ {s \in N(m)} p _ {uis} ^ h c_s.
\end{equation}
Here, $\mathbf{o} _ {ui} ^ h$ is the representation vector of snapshot models after considering the neighborhood representation propagated from previous hops, which is then transferred to the next hop as follows:
\begin{equation}
\mathbf{z} _ {u i} ^ h = \phi (\mathbf{W} ^ h \mathbf{z} _ {u i} ^ {h - 1} + \mathbf{o} _ {u i} ^ h + \mathbf{b} ^ h),
\end{equation}
where the projection matrix $\mathbf{W} ^ h \in \mathbb{R} ^ {D \times D}$ and the bias vector $\mathbf{b} ^ h \in \mathbb{R} ^ D$ are parameters. $\phi$ denotes the activation function and $\mathbf{z} _ {u i} ^ {h - 1}$ represents the input query of hop $h$. The initial query $z _ {u i} ^ 0 = \mathbf{e} _ {u i}$. In the last hop, the fully connected layer produces the final prediction $\hat {\mathbf{y}} \in \mathbb{R} ^ {N_m}$ as ensemble weights, which is the same as the single hop memory network.

\subsection{Loss Function with Soft Labels}

Our goal is to combine snapshot models not to select the optimal one because the optimal snapshot model is risky to infer due to the limited ratings of individual users/items, so that combining snapshot models whose training epochs are close to optimum can help to reduce ensemble variance. Therefore, we formulate the ensemble learning task as a multi-class classification problem. More specifically, we propose a soft labelling method to assign larger weights for snapshot models whose training epochs are close to optimum. The soft label for snapshot model $s$ is defined as follows:
\begin{equation}
y_s = \frac {\exp(x_s)} {\sum _ {k \in N(m)} \exp(x_k)}, \quad \forall s \in N(m). 
\end{equation}
Here, $x_s =  (| e_s - e_o| + 1) ^ {-\alpha}\in(0,1]$ is the transformed epoch distance between the $s^{th}$ snapshot model ($e_s$) and the optimal one ($e_o$). And $\alpha$ is a hyperparameter, which is chosen as 1 based on empirical studies.

To measure the difference between the predicted ensemble weights distribution $\hat {\mathbf{y}}$ and the targeted distribution $\mathbf{y}$ defined above, we adopt the KL-divergence described as follows:
\begin{equation}
\textstyle
L = \sum _ {s = 1} ^ {N_m} y_s (\log y _ s - \log \hat y _ s).
\end{equation}
This is also the loss function to train the ensemble network, which can be effectively solved via SGD. After obtaining the ensemble weights for each user-item pair (from the output layer of the proposed ensemble network), we can use a simple weighted average to obtain the recommendation scores as follows:
\begin{equation}
\tilde{r}_{ui} = \hat{\mathbf{y}}\cdot\hat{\mathbf r}_{ui}.
\end{equation}
$\hat{\mathbf{y}}$ is the predicted ensemble weights for the targeted user-item pair, and $\hat{\mathbf r}_{ui}$ contains the predicted scores on the targeted user-item pair from all snapshot models.

\section{Discussion}

\subsection{Efficiency Analysis}
The computational overhead of training the NeuSE models is small compared to training many collaborative filtering models, \textit{e.g.}, NeuMF~\cite{NCF}. In each epoch, the computational complexity is $O(DHN_R)$, where $D$ is the embedding size of the memory networks, $H$ is the number of hops of the memory networks and $N_R$ is the number of ratings. Due to the adoption of multiple memory network components, the proposed method requires higher memory space in model training compared with NeuMF. However, as shown in the experiments, although NeuSE may take more memory space compared with NeuMF, the overall training time of NeuSE is much lower than that of NeuMF due to much faster convergence. More empirical discussions about the efficiency of the proposed method can be found in the experiments.

\subsection{Addressing Cold-start User/item}
The cold-start problem is a well-known challenge in collaborative filtering~\cite{survey05}. The proposed method can handle cold-start user or cold-start item as follows. We can change the input layer from the simple one-hot vectors to the user/item rating vectors, then the proposed method can handle unseen users who have ratings on seen items (cold-start user problem) or unseen items who have ratings by seen users (cold-start item problem). But for unseen users who only rate unseen items, we may need to either re-train the models or rely on auxiliary information (e.g., user demographics or item genres) as other CF algorithms.

\subsection{Application Scope}
Existing collaborative filtering algorithms can be divided into two categories: (1) 
non-learning based methods (e.g., most popular and  KNN~\cite{Herlocker99,Sarwar01}) and learning based methods (e.g.,  factorization methods~\cite{koren2009matrix,BPMF,FM} and neural network methods~\cite{NCF,Sedhain15}). We now analyze the feasibility of the proposed NeuSE method on top of each kind of methods.

In non-learning based methods, if we cannot create snapshot models, e.g., in most popular-based recommendation, the proposed NeuSE method will fail to improve the performance. However, if we can create snapshot models, e.g., by choosing different number of neighbors in KNN methods, NeuSE can successfully improve the performance as demonstrated in the experiments (Section~\ref{sec:exp}).

In learning based methods, we can analyze the feasibility of NeuSE according to the optimization problems of the collaborative filtering algorithms. In most learning based methods, the model learning problems are formulated as non-convex optimization problems, e.g., as defined in Equation~\ref{eqn:loss}, in which iterative methods, e.g., SGD, are popular ways to learn the models. In this case, we can easily obtain snapshot models by choosing the intermediate models during learning. However, if the model learning problems are formulated as convex optimization problems~\cite{aiolli2014convex,steck2019embarrassingly}, the learned model can be obtained by a closed form solution. In this case, we cannot apply NeuSE to improve their performances because we have no snapshot models. However, convex optimization is rare in collaborative filtering algorithms, so the we can claim that NeuSE can be applied to most of the  learning-based collaborative filtering algorithms.

In summary, NeuSE can be applied to improve the model performance as long as we can create snapshot models. As demonstrated in the experiments, snapshot models can be created in most of the popular collaborative filtering algorithms, so that we believe NeuSE is desirable in collaborative filtering tasks.

\section{Experiments}
\label{sec:exp}
\subsection{Experimental Setup}
\label{sec:exp_setup}
\begin{table}[tb!]
	\caption{Statistics of the datasets adopted in the experiments.}
	\label{tab:data}
	\small
	\begin{tabular}{c|c|c|c|c}
		\hline
		Dataset & \#users & \#items & \#ratings & Density \\
		\hline
		MovieLens 100K\footnote{\url{https://grouplens.org/datasets/movielens/100k/}}
		 & 943 & 1,682 & 100,000 & 6.30\%\\
		\hline
		Book Crossing\footnote{\url{http://www2.informatik.uni-freiburg.de/~cziegler/BX/}}
		 & 2,946  & 17,384 & 272,679 & 0.53\%\\
		\hline
		Amazon Musical Instruments\footnote{\url{http://jmcauley.ucsd.edu/data/amazon/}}
		 & 80,004 & 13,836 & 500,176 & 0.05\%\\
		\hline
	\end{tabular}
\end{table}

\subsubsection{Dataset Description.}
We adopt three popular real-world datasets to evaluate the performance of the proposed snapshot ensemble method, the details of which are listed in Table~\ref{tab:data}. All three datasets can be regarded as both explicit feedback (user ratings on items) and implicit feedback (ratings as positive feedbacks and others as unknown feedbacks). The rating scale is 1-5 in the MovieLens and Amazon datasets and 1-10 in Book Crossing dataset.

\subsubsection{Evaluation Protocol.}
For all experiments, we use the leave-one-out evaluation protocol~\cite{NCF,Rendle09}. We split the datasets into training, validation and test sets by chronological order, in which the most recent rating of each user is used in test, the penultimate rating is used in validation and all the rest ratings are used in training. Then, the best performing models on the validation set are selected for test. For top-N recommendation, we randomly select 99 negative examples for each test example~\cite{NCF}. The negative examples are then fixed in the same experiment for fair comparison.

\subsubsection{Compared Methods.}
We compare the proposed snapshot ensemble method with the following methods:
\begin{itemize}
\item {\bf Single}: the single snapshot model with best validation performance is used for prediction;
\item {\bf Average}: the horizontal averaging method~\cite{horizontal}, which averages the outputs of snapshot models for prediction;
\item {\bf HSE}: the horizontal stacked ensemble method~\cite{horizontal}, which takes the outputs of snapshot models as the inputs to another meta model. We use an MLP as the meta learner for HSE in the experiments due to the universal approximation ability of deep neural networks.
\item {\bf SE}: the cyclic learning rate-based	snapshot ensemble method~\cite{snapshot}, which is a state-of-the-art snapshot ensemble method but only applicable in models learned via SGD.
\end{itemize}

\subsubsection{Baseline CF Methods.}
For each snapshot ensemble method, we apply them on the following popular collaborative filtering algorithms:
1) {\bf UserKNN}~\cite{Herlocker99};
2) {\bf ItemKNN}~\cite{Sarwar01};
3) {\bf RSVD}~\cite{paterek2007improving};
4) {\bf FM}~\cite{FM};
5) {\bf AutoRec}~\cite{Sedhain15};
6) {\bf NeuMF}~\cite{NCF}.
The above baselines cover many kinds of existing CF methods including three algorithm types (memory-based, factorization, and neural network methods) and two tasks (rating prediction and top-N recommendation).

\subsubsection{Hyperparameters.}
In training the proposed NeuSE network, we use the batch size of 128, initial learning rate of 0.01, and Adam~\cite{kingma2014adam} as the optimizer to adaptively tune the learning rate. We apply dropout~\cite{Dropout} in NeuSE with a dropout rate of 0.5. We use ReLU as the activation function for NeuSE due to better empirical performance in deep models~\cite{ReLU}. The user/item embedding size of NeuSE is set to 16 for all the experiments. We randomly initialize the parameters of NeuSE with a Gaussian distribution $\mathcal N(0, 0.01)$. We use the two-hop memory network architecture in the experiments unless otherwise specified.

The hyperparameters of all baseline CF methods are kept the same for all snapshot ensemble methods as reported in their original papers, \textit{i.e.}, the performances are only related to the snapshot ensemble methods. For RSVD, FM and AutoRec, we use the epoch variations $\Delta T = 10$ to select snapshot models and $\Delta T=5$ for NeuMF till 200 epochs. For UserKNN and ItemKNN, we use cosine similarity to find nearest neighbors and use the number of nearest neighbors in $\{10,20,30,...,100\}$ to create snapshot models. The embedding sizes or number of hidden factors are 8, 8, 500, 32 for RSVD, FM, AutoRec and NeuMF, respectively. Note that NeuSE can also improve the performance of baseline CF algorithms with large embedding sizes as demonstrated in the sensitivity analysis. Other hyperparameters of the baseline CF algorithms, e.g., learning rates, weight decay, etc., are adopted from the best-performing ones in the original papers.

\subsubsection{Evaluation Metrics.}
For rating prediction task, we evaluate the performances of different methods using root mean square error (RMSE), in which lower value indicates better performance:
\begin{displaymath}
\mathrm{RMSE} = \sqrt{\frac{1}{|\Omega|}\sum_{u,i\in\Omega}(R_{u,i}-\hat{R}_{u,i})^2},
\end{displaymath}
where $R_{u,i}$ is the rating of user $u$ on item $i$ and $\hat{R}_{u,i}$ is the corresponding predicted rating. $\Omega$ contains all the user-item pairs in the test set and $|\Omega|$ denotes the number of test examples.

For top-N recommendation task, we evaluate the performances of different methods using two popular metrics: Hit Ratio@N (HR@N) and NDCG@N, in which higher value indicates better performance:
\begin{eqnarray}
& \mathrm{HR@N} = \frac{\mathrm{\#Users~w.~Hits~@N}}{\mathrm{\#Users}},  \\
& \mathrm{NDCG@N} = \frac{\mathrm{DCG@N}}{\mathrm{IDCG@N}}. 
\end{eqnarray}
Here, $\mathrm{DCG@N} = \sum_{i=1}^N\frac{2^{r_i}-1}{\log_2(i+1)}$, where $r_i$ is 1 if the $i$-th item in the top N list is relevant to the targeted user and 0 otherwise. $\mathrm{IDCG@N}$ is the value of $\mathrm{DCG@N}$ with perfect ranking.

\begin{table*}[t!]
	\centering
	\caption{Accuracy comparison between NeuSE and four snapshot ensemble methods (Single, Average~\cite{horizontal}, HSE~\cite{horizontal} and SE~\cite{snapshot}) on the MovieLens 100K, Book Crossing and Amazon Musical Instruments datasets. Note that the SE method can only be applied to models learned via SGD and thus cannot be applied to  UserKNN and ItemKNN. For the AutoRec method, we report the results of item-based AutoRec (I-AutoRec), which showed better performance than the U-AutoRec method in their paper. $N$ is set to 20 for HR@N and NDCG@N in the results.
		\label{tab:accuracy}}
	\small
	\begin{tabular}{c|c|ccc|ccc|ccc}
		\hline
		& \multirow{2}{*}{{Method}}  & \multicolumn{3}{c|}{{MovieLens 100K}} & \multicolumn{3}{c|}{{Book Crossing}} & \multicolumn{3}{c}{{Amazon}}  \\
		\cline{3-11}
		& & {RMSE} & {HR} & {NDCG} & {RMSE}& {HR}& {NDCG} & {RMSE} & {HR}& {NDCG} \\
		\hline
		\multirow{4}{*}{UserKNN}
		& Single 	  & 1.2727 & 0.2365 & 0.0822 & 2.2056 & 0.2423 & 0.0844 & 1.3126 & 0.1548 & 0.0481 \\
		& Average     & 1.2759 & 0.2354 & 0.0801 & 2.2156 & 0.2410 & 0.0841 & 1.3118 & 0.1538 & 0.0473 \\
		& HSE		  & 1.2704 & 0.2384 & 0.0827 & 2.2124 & 0.2415 & 0.0865 & 1.3142 & 0.1572 & 0.0446\\
		& {\bf NeuSE} & \textbf{1.2615} & \textbf{0.2398} & \textbf{0.0857} & \textbf{2.1847} & \textbf{0.2562} & \textbf{0.0897} & \textbf{1.3046} & \textbf{0.1597} & \textbf{0.0517} \\
		\hline
		\multirow{4}{*}{ItemKNN}
		& Single      & 1.2755 & 0.2238 & 0.0811 & 2.2152 & 0.2289 & 0.0791 & 1.3061 & 0.2257 & 0.0851 \\
		& Average     & 1.2753 & 0.2216 & 0.0802 & 2.2156 & 0.2241 & 0.0725 & 1.3085 & 0.2259 & 0.0843 \\
		& HSE		  & 1.2682 & 0.2269 & 0.0812 & 2.2065 & 0.2292 & 0.0738 & 1.2957 & 0.2234 & 0.0845 \\
		& {\bf NeuSE}& \textbf{1.2621} & \textbf{0.2306} & \textbf{0.0862} & \textbf{2.1828} & \textbf{0.2395} & \textbf{0.0837} & \textbf{1.2879} & \textbf{0.2343} & \textbf{0.0886} \\
		\hline
		\multirow{5}{*}{RSVD}
		& Single         & 1.1305 & - & - & 2.2849 & - & - & 1.2851 & - & - \\
		& Average     & 1.1171 & - & - & 2.2762 & - & - & 1.2831 & - & - \\
		& HSE		  & 1.1327 & - & - & 2.2851 & - & - & 1.2846 & - & - \\
		& SE             & 1.1130 & - & - & 2.2758 & - & - & 1.2818 & - & - \\
		& {\bf NeuSE}& { \textbf{1.0973}} & - & - & { \textbf{2.2686}} & - & - & {\textbf{1.2689}} & - & - \\
		\hline
		\multirow{5}{*}{\makecell{FM\\(SGD)}}
		& Single      & 1.1449 & 0.6278 & 0.2779 & 2.2351 & 0.3741 & 0.1403 & 1.1136 & 0.4759 & 0.1762 \\
		& Average     & 1.0948 & 0.6215 & 0.2631 & 2.1980 & 0.3688 & 0.1396 & 1.1052 & 0.4758 & 0.1658 \\
		& HSE		  & 1.1047 & 0.6324 & 0.2791 & 2.2031 & 0.3764 & 0.1438 & 1.1037 & 0.4771 & 0.1784 \\
		& SE          & 1.0942 & 0.6307 & 0.2759 & 2.1975 & 0.3798 & 0.1457 & 1.1028 & 0.4789 & 0.1777 \\
		& {\bf NeuSE} & {\bf{1.0928}} & {\bf 0.6382} & {\bf 0.2893} & {\bf{2.1938}} & {\bf 0.3826} & {\bf 0.1497}
		& {\bf{1.0970}} & {\bf 0.4844} & {\bf 0.1813} \\
		\hline
		\multirow{4}{*}{\makecell{FM\\(MCMC)}}
		& Single      & 1.4138 & 0.5882 & 0.2631 & 2.5923 & 0.3672 & 0.1285 & 1.5391 & 0.4577 & 0.1525 \\
		& Average     & 1.4157 & 0.5895 & 0.2674 & 2.5892 & 0.3668 & 0.1273 & 1.5348 & 0.4542 & 0.1487 \\
		& HSE		  & 1.4154 & 0.5841 & 0.2609 & 2.5854 & 0.3654 & 0.1274 & 1.5301 & 0.4594 & 0.1549 \\
		& {\bf NeuSE} & {\bf{1.3812}} & {\bf 0.5977} & {\bf 0.2704} & {\bf{2.4647}} & {\bf 0.3713} & {\bf 0.1324}
		& {\bf{1.4954}} & {\bf 0.4643} & {\bf 0.1586} \\
		\hline
		\multirow{5}{*}{I-AutoRec}
		& Single		& 1.1078 & 0.6776 & 0.2983 & 1.6479 & 0.3859 & 0.1427 & 0.9552 & 0.5127 & 0.2359 \\
		& Average		& 1.1164 & 0.6752 & 0.2985 & 1.6482 & 0.3814 & 0.1385 & 0.9517 & 0.5115 & 0.2381 \\
		& HSE			& 1.1029 & 0.6739 & 0.2901 & 1.6483 & 0.3847 & 0.1417 & 0.9537 & 0.5163 & 0.2379 \\
		& SE			& 1.1084 & 0.6734 & 0.2939 & 1.6375 & 0.3879 & 0.1462 & 0.9586 & 0.5129 & 0.2367 \\
		& {\bf NeuSE}& \textbf{1.0857} & \textbf{0.6845} & \textbf{0.3017} & \textbf{1.5849} & \textbf{0.3924} & \textbf{0.1489} & \textbf{0.9348} & \textbf{0.5231} & \textbf{0.2427} \\
		\hline
		\multirow{4}{*}{NeuMF}
		& Single        & - & 0.8134 & 0.4087 & - & 0.4423 & 0.1948 & - & 0.6750 & 0.4083 \\
		& Average       & - & 0.8081 & 0.4097 & - & 0.4532 & 0.1943 & - & 0.6752 & 0.4085 \\
		& HSE			& - & 0.8092 & 0.4075 & - & 0.4527 & 0.1963 & - & 0.6784 & 0.4098 \\
		& SE            & - & 0.8171 & 0.4121 & - & 0.4535 & 0.1952 & - & 0.6755 &  0.4083\\
		& {\bf NeuSE}& - & {\textbf{0.8297}} & { \textbf{0.4153}} & - & {\textbf{0.4679}} & { \textbf{0.2031}} & - & \textbf{0.6847} & \textbf{0.4124} \\
		\hline
	\end{tabular}
\end{table*}

\subsection{Recommendation Accuracy}
Table~\ref{tab:accuracy} compares the accuracy between NeuSE and three snapshot ensemble methods on the MovieLens, Book Crossing and Amazon datasets. As shown in the table, the accuracy gain of NeuSE varies across different baseline CF methods and datasets. The highest improvements are $\sim$7.5\% over single model at the NDCG of UserKNN in the Amazon dataset, $\sim$15.4\% over Average method at the NDCG of ItemKNN in the Book Crossing dataset, $\sim$15.9\% over the HSE method at the NDCG of UserKNN in the Amazon dataset and $\sim$4.1\% over the SE method at the NDCG of NeuMF in the Book Crossing dataset. Although the improvements are not so significant everywhere, we can see that the improvements are universal, \textit{i.e.}, over all baseline algorithms on all datasets.

We have the following observations from the results:
\begin{itemize}
	\item Simply averaging the snapshot models cannot always improve accuracy over the optimal snapshot model, in which simple average increases the accuracy in three out of six baseline algorithms. The main reason is that appropriate subset of models could be better than all of them~\cite{Zhou02}, but simple average takes all snapshot models with equal weights and thus cannot alleviate the effects of poor snapshot models.
	\item The HSE method, which uses an MLP to combine the snapshot models, achieves better performance than simple average in most of cases. This further confirms that dynamic ensemble (by varying the ensemble weights for different examples) could be more desirable in collaborative filtering tasks. However, the HSE method cannot memorize the user- and item-specific characteristics, e.g., some users/items may converge early as shown in the case studies, and thus cannot achieve optimal accuracy.
	\item The SE method using cyclic learning rate to obtain snapshot models achieves comparable accuracy to simple average on factorization methods (i.e., RSVD and FM) and slightly better accuracy on neural network models (i.e., AutoRec and NeuMF). The main reason is that neural network models trained with cyclic learning rate may converge to different local optimums~\cite{snapshot} but factorization models may converge to the same local optimum if using cyclic learning rate.
	\item The proposed NeuSE method can achieve higher accuracy than all other three kinds of snapshot ensemble methods, because a) NeuSE adaptively weighs different snapshot models for individual ratings so that the effects of poor models can be alleviated from the ensemble; b) different snapshot models can converge to different local optimums due to the nature of iterative optimization; c) the proposed ensemble method can fully utilize the global importance of snapshot models, the user and item specific characteristics and the neighborhood information from similar users and items via the memory networks to achieve better performance.
	\item We can also observe that the proposed NeuSE method can improve the accuracy on models trained with both iterative optimization methods (e.g., SGD) and Bayesian approaches (e.g., MCMC). In addition, the proposed method can also improve the accuracy on memory-based methods (e.g., UserKNN and ItemKNN), in which no learning procedure is involved. This indicates that NeuSE can be applied to almost all kinds of existing collaborative filtering algorithms to boost their accuracy.
\end{itemize}

\subsection{Sensitivity Analysis}
We first evaluate the sensitivity of NeuSE with different memory layers and labels and then analyze its efficiency. Note that we only report the results on three CF methods: RSVD, FM and NeuMF on the MovieLens 100K dataset because other methods and datasets show similar trends and thus are omitted.

\begin{figure*}[tb!]
	\centering
	\includegraphics[width=0.325\textwidth]{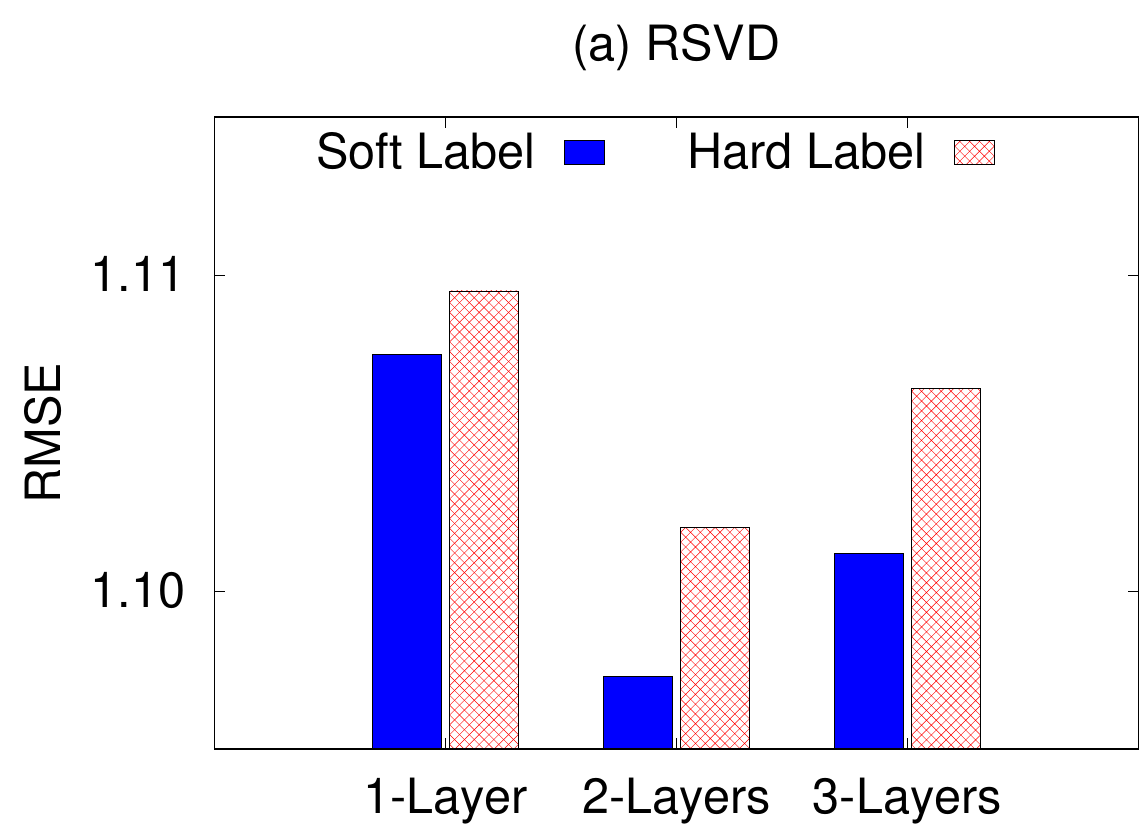}
	\includegraphics[width=0.325\textwidth]{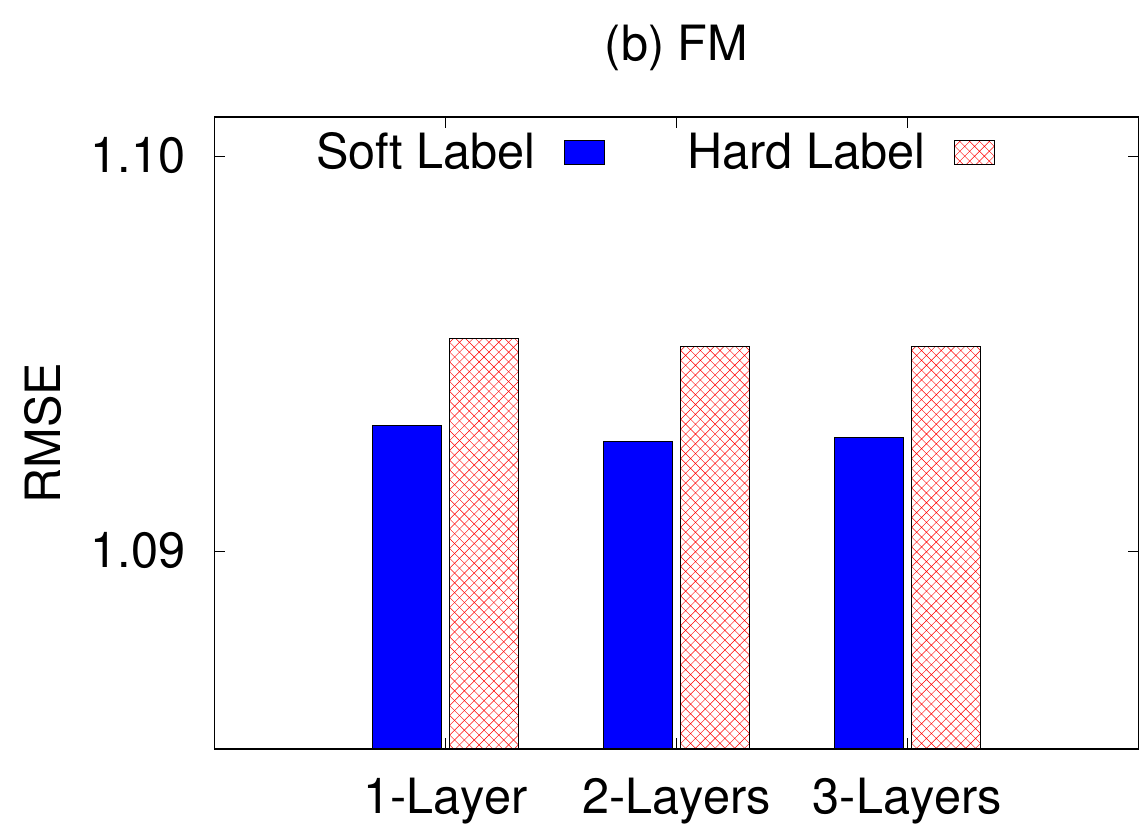}
	\includegraphics[width=0.325\textwidth]{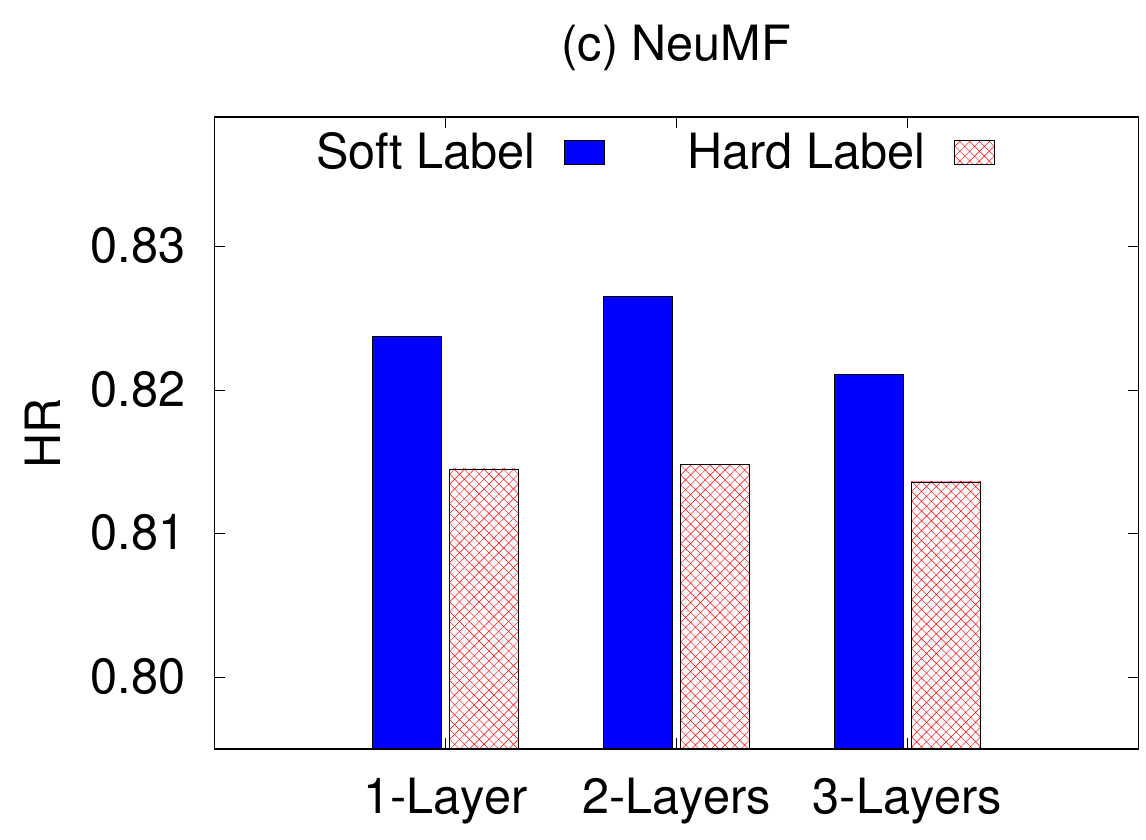}
	\caption{Sensitivity of NeuSE with the number of memory network layers and labels on three CF algorithms: (a) RSVD, (b) FM and (c) NeuMF. We use ReLU as the activation function and 16 as embedding size in this experiment.
		\label{fig:sensitivity}}
\end{figure*}

\subsubsection{Memory Network Layers} This experiment analyzes how the accuracy of NeuSE changes with the number of memory layers. As shown in Figure~\ref{fig:sensitivity}, for all three CF methods, two-layer memory networks achieve the highest accuracy than one-layer and three-layer memory networks. The main reason is that one-layer memory network cannot leverage the global/neighborhood information in previous layers to infer better representations in current layer and three-layer memory network is prone to overfitting due to more parameters.

\subsubsection{Soft vs. Hard Label} This experiment analyzes how the proposed soft labelling technique affects the accuracy of NeuSE. As shown in Figure~\ref{fig:sensitivity}, NeuSE learned with soft labels achieves higher accuracy over hard labels, $\sim$0.5\% improvement in RSVD, $\sim$0.2\% improvement in FM, and $\sim$1.4\% improvement in NeuMF. This confirms that the proposed soft-labelling method can indeed improve the performance of snapshot ensemble due to the ability of adapting the losses on snapshot models with different performances. On the contrary, hard labels are prone to overfitting due to the noisy behaviors of individual users/items.

\begin{figure*}[tb!]
	\centering
	\includegraphics[width=0.325\textwidth]{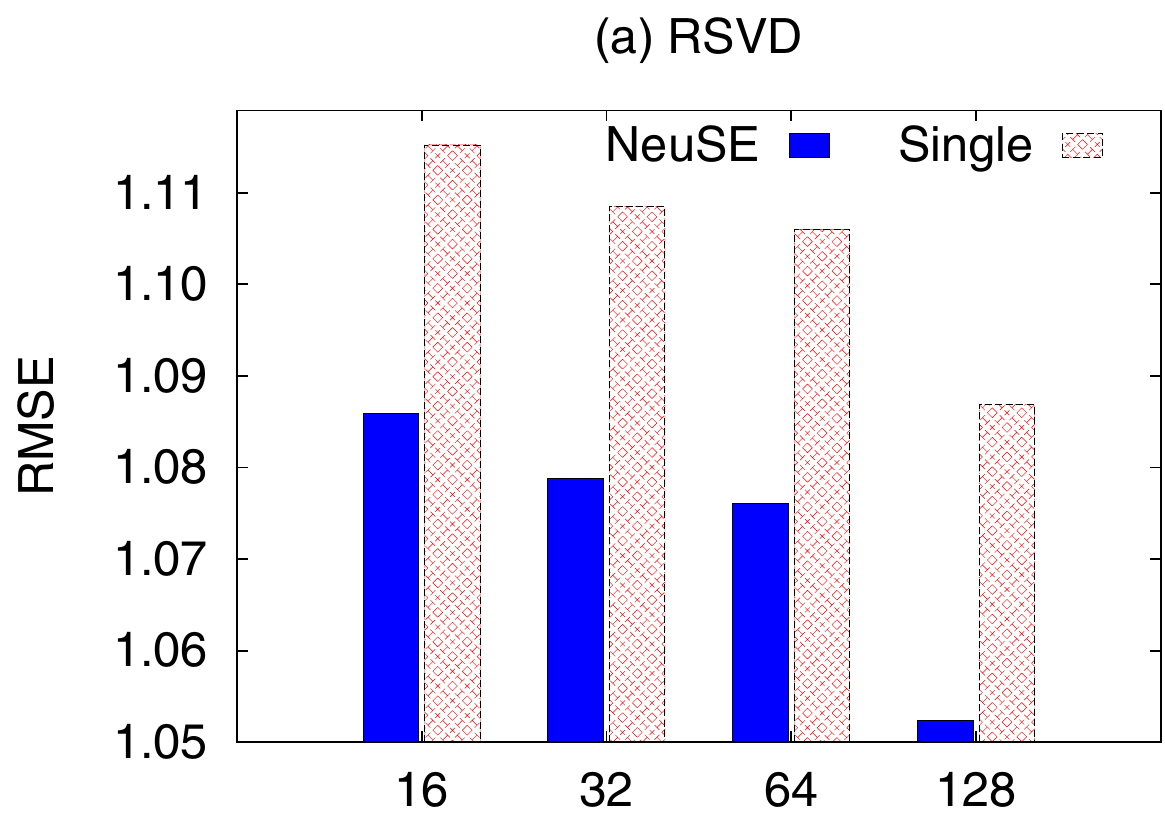}
	\includegraphics[width=0.325\textwidth]{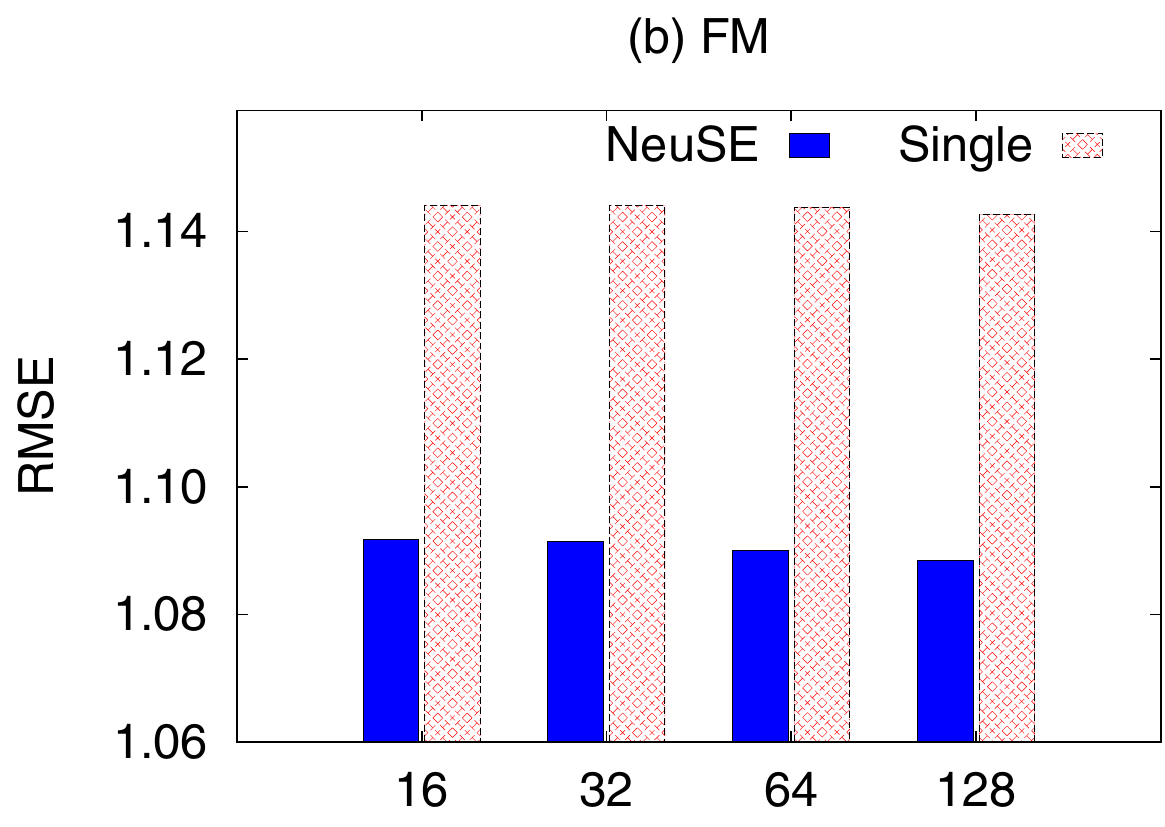}
	\includegraphics[width=0.325\textwidth]{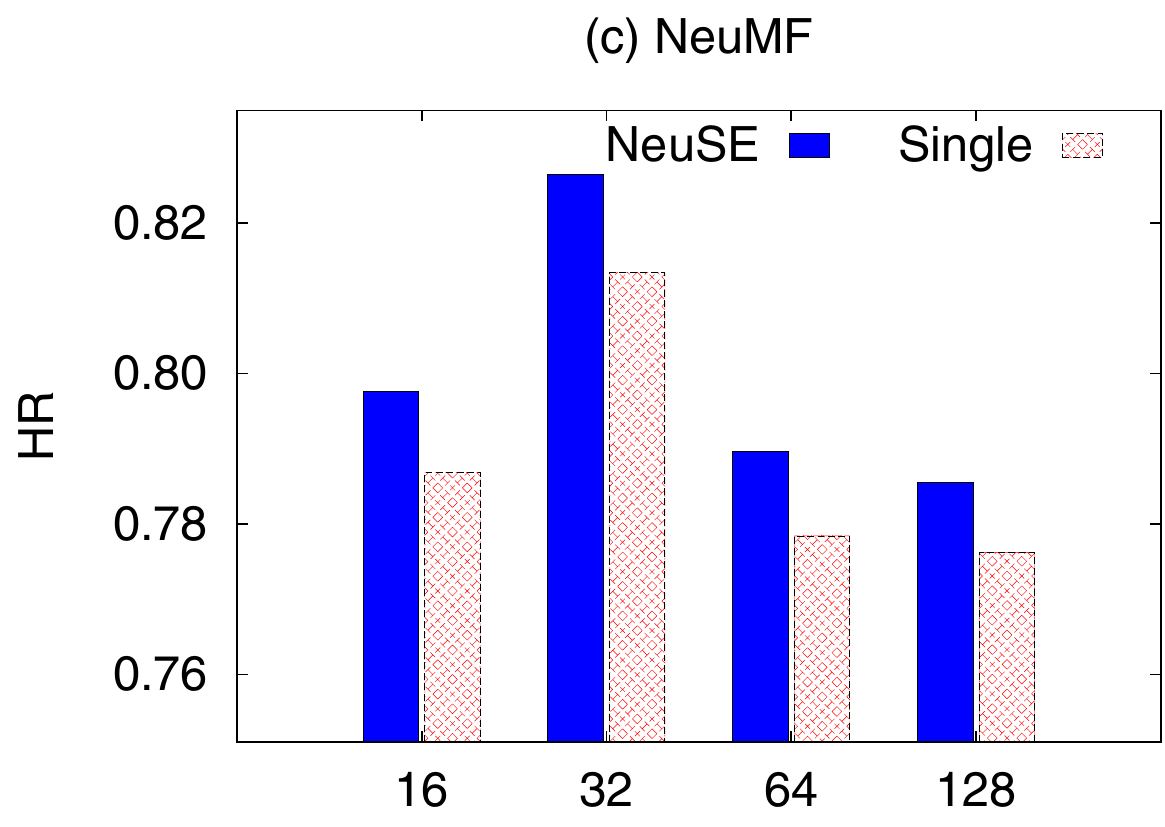}
	\caption{Sensitivity of NeuSE with CF models of different complexity on three CF algorithms: (a) RSVD, (b) FM and (c) NeuMF. We use ReLU as the activation function in this experiment. The $x$-axis denotes the complexity (number of latent factors and embedding size) of the CF models in each algorithm.
		\label{fig:complexity}}
\end{figure*}

\subsubsection{Model Complexity}
Due to the possible concern that NeuSE can only improve the performance of baseline CF algorithms with low capacity, we conduct experiments to analyze the performance of NeuSE in CF models with different model complexity, e.g., different user/item embedding sizes in NeuMF. As shown in Figure~\ref{fig:complexity}, NeuSE can consistently improve the performance over the best Single models in all the cases with different complexities of the underlying CF models. This experiment indicates that (1) the performance of CF algorithms will vary with model capacity but the suboptimal performance issue exists in both high capacity models and low capacity models and (2) NeuSE can consistently improve the performance as long as the suboptimal performance issue exists.

\begin{figure*}[tb!]
	\centering
	\includegraphics[width=0.325\textwidth]{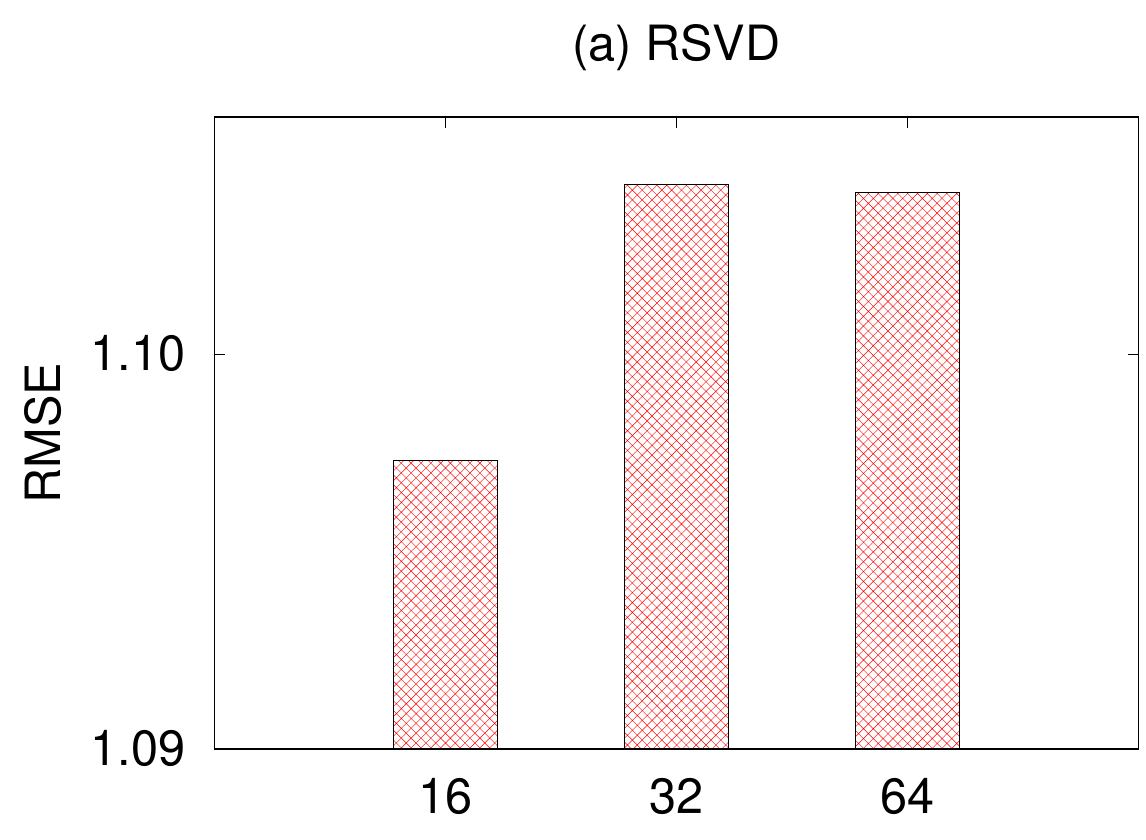}
	\includegraphics[width=0.325\textwidth]{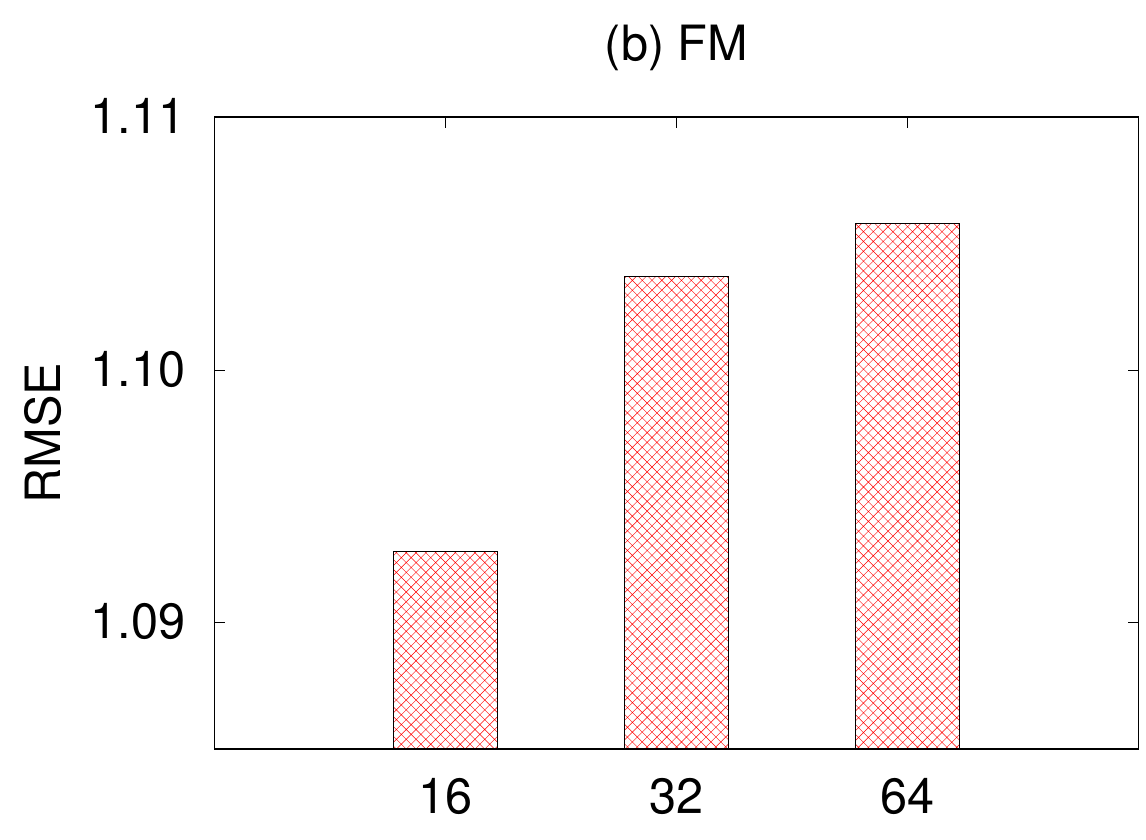}
	\includegraphics[width=0.325\textwidth]{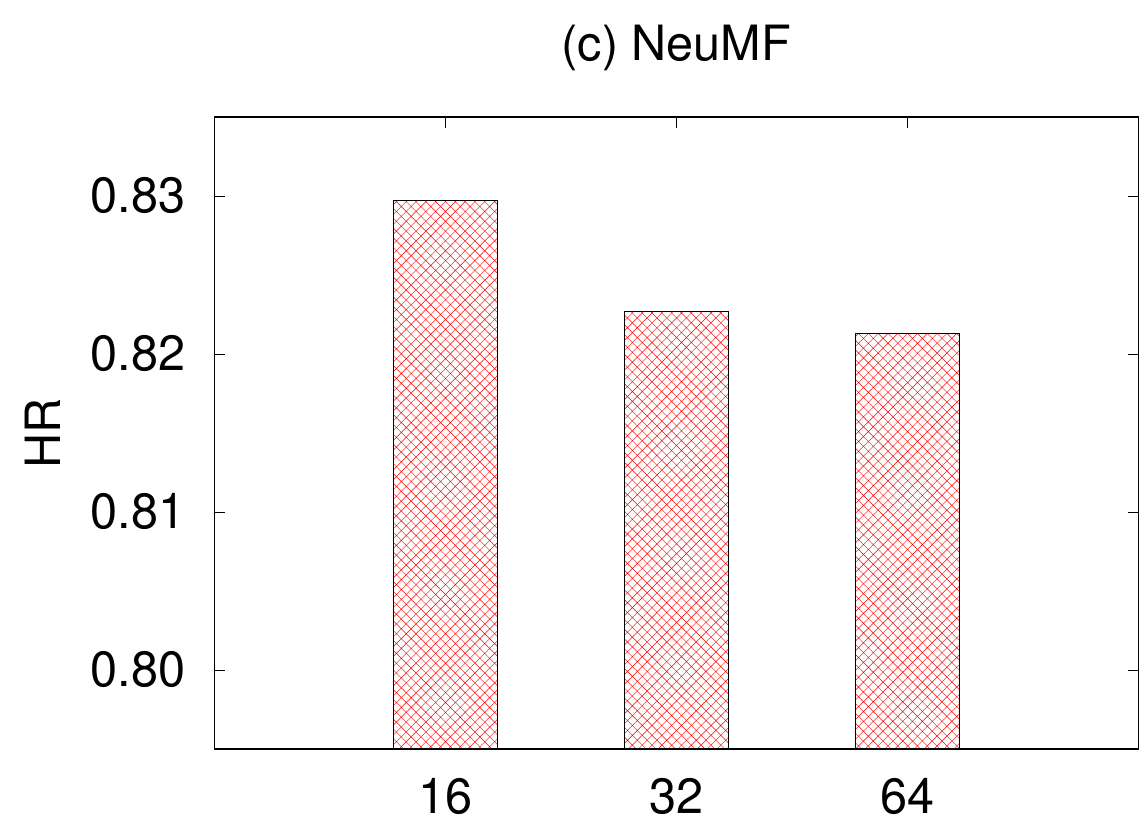}
	\caption{Sensitivity of NeuSE with different user/item embedding sizes (16, 32, 64) on three CF algorithms: (a) RSVD, (b) FM and (c) NeuMF. We use ReLU as the activation function in this experiment.
		\label{fig:embedding}}
\end{figure*}

\subsubsection{Embedding Size}
This experiment analyzes how sensitive are NeuSE models to different sizes of user/item embedding. As shown in Figure~\ref{fig:embedding}, the accuracy of NeuSE peaks with the embedding size of 16 and the sign of overfitting appears when embedding size grows to 32 and 64. This  indicates that NeuSE can achieve decent performance with very light-weight models due to the adoption of multi-layer memory networks. Small embedding size also indicates that training NeuSE models can enjoy fast computation and high memory efficiency, i.e., NeuSE is less likely to suffer from the efficiency bottlenecks compared with many existing ensemble collaborative filtering methods.

\begin{figure*}[tb!]
	\centering
	\includegraphics[width=0.325\textwidth]{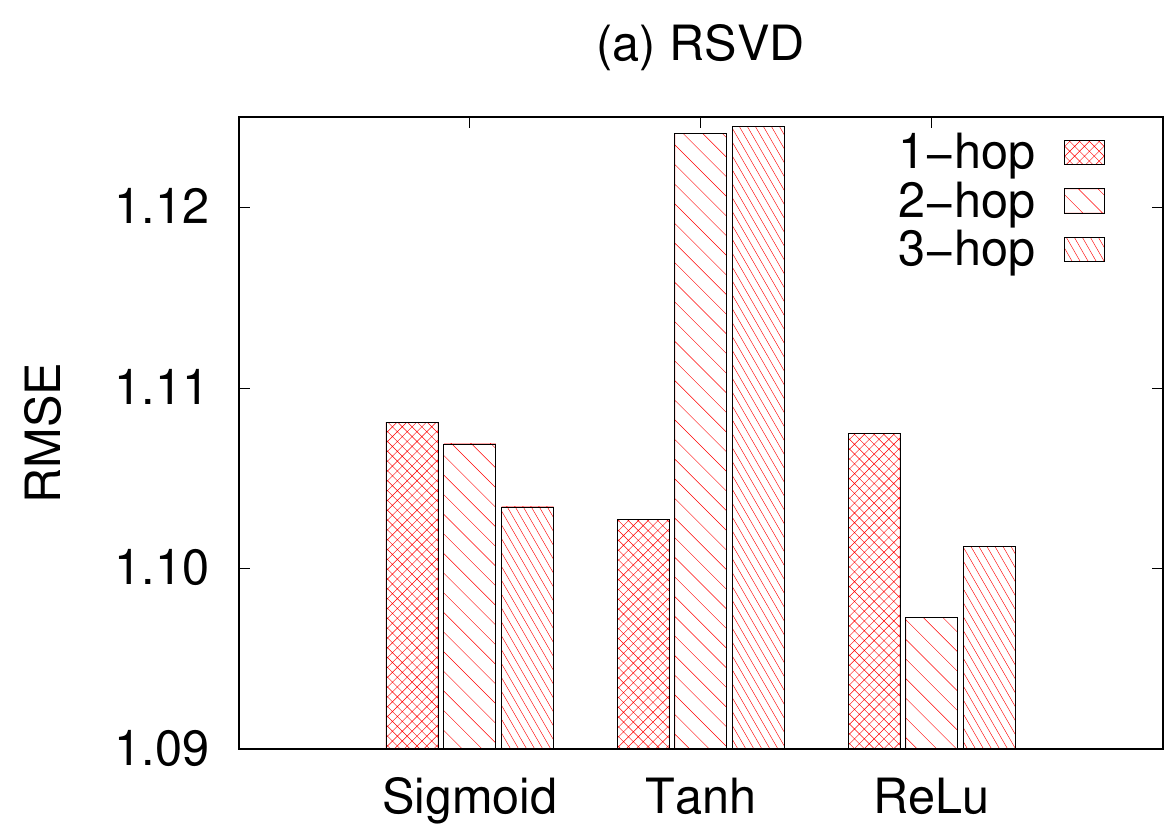}
	\includegraphics[width=0.325\textwidth]{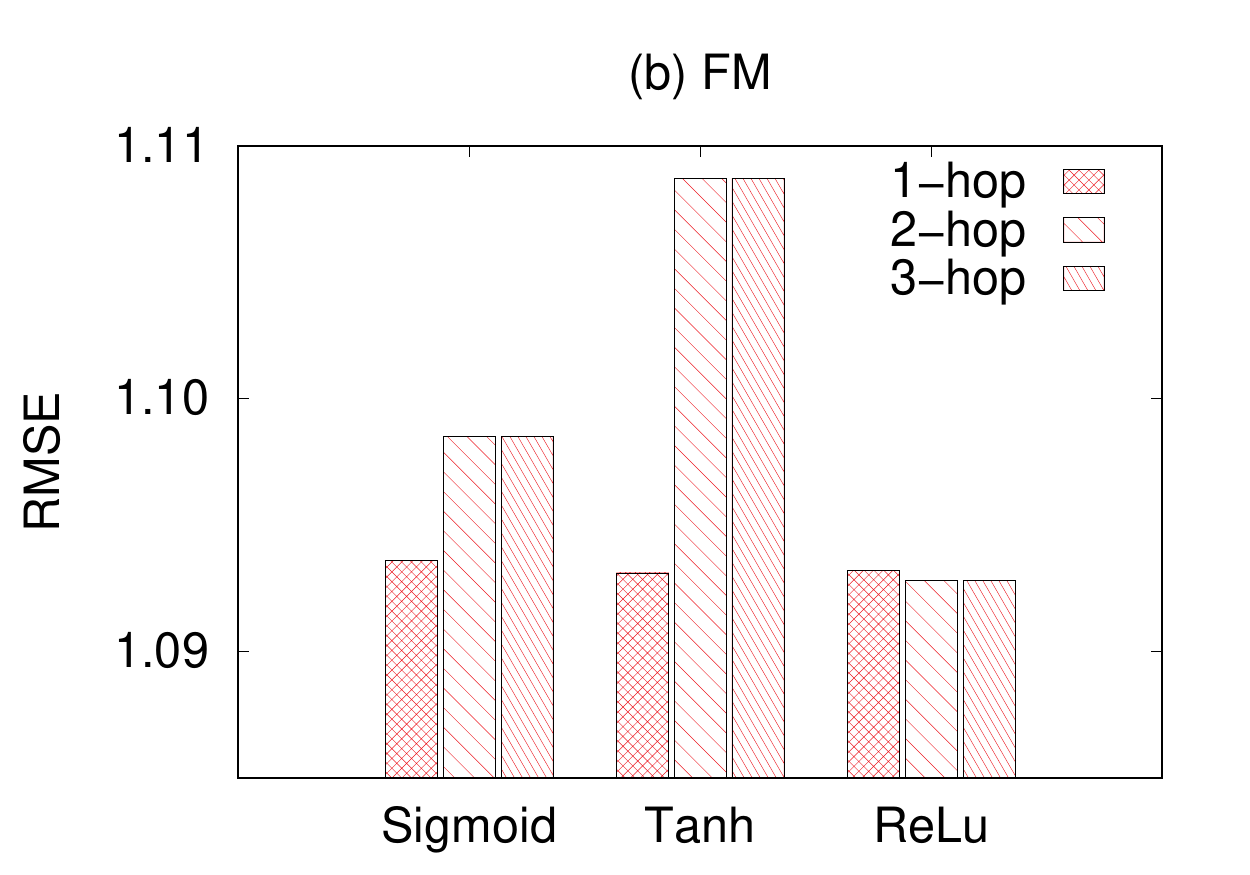}
	\includegraphics[width=0.325\textwidth]{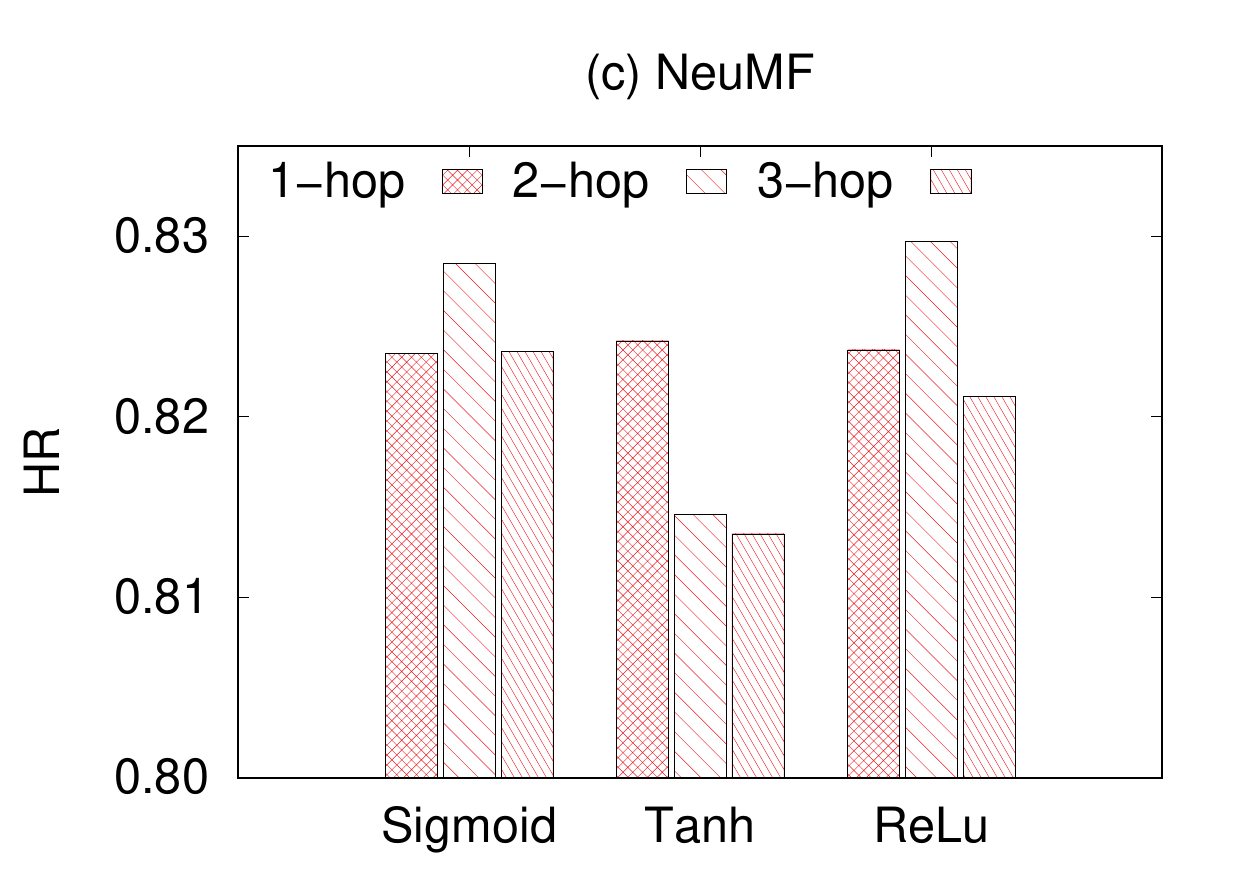}
	\caption{Sensitivity of NeuSE with different activation functions (Sigmoid, Tanh, ReLU) on three CF algorithms: (a) RSVD, (b) FM and (c) NeuMF. We use the embedding size of 16 in this experiment. On average, ReLU achieves the best performance among all three activation functions.
		\label{fig:activation}}
\end{figure*}

\subsubsection{Activation Function}
This experiment analyzes how sensitive is NeuSE to different activation functions in the hidden layers. As shown in Figure~\ref{fig:activation}, all three activation functions achieve similar performance if we only use one hop of memory network. When we increase the  memory networks to 2-hop or 3-hop, Tanh achieves the worst performance among all three activation functions. The performance of ReLU is slightly better than Sigmoid and Tanh on average in all three algorithms.  Compared with Sigmoid and Tanh, ReLU can reduce the likelihood of gradient vanishing and achieve better convergence in practice~\cite{krizhevsky2012imagenet}, which could be the main reasons of the better performance of ReLU.

\begin{figure*}[tb!]
	\begin{minipage}[t]{0.32\linewidth}
		\includegraphics[width=1\textwidth]{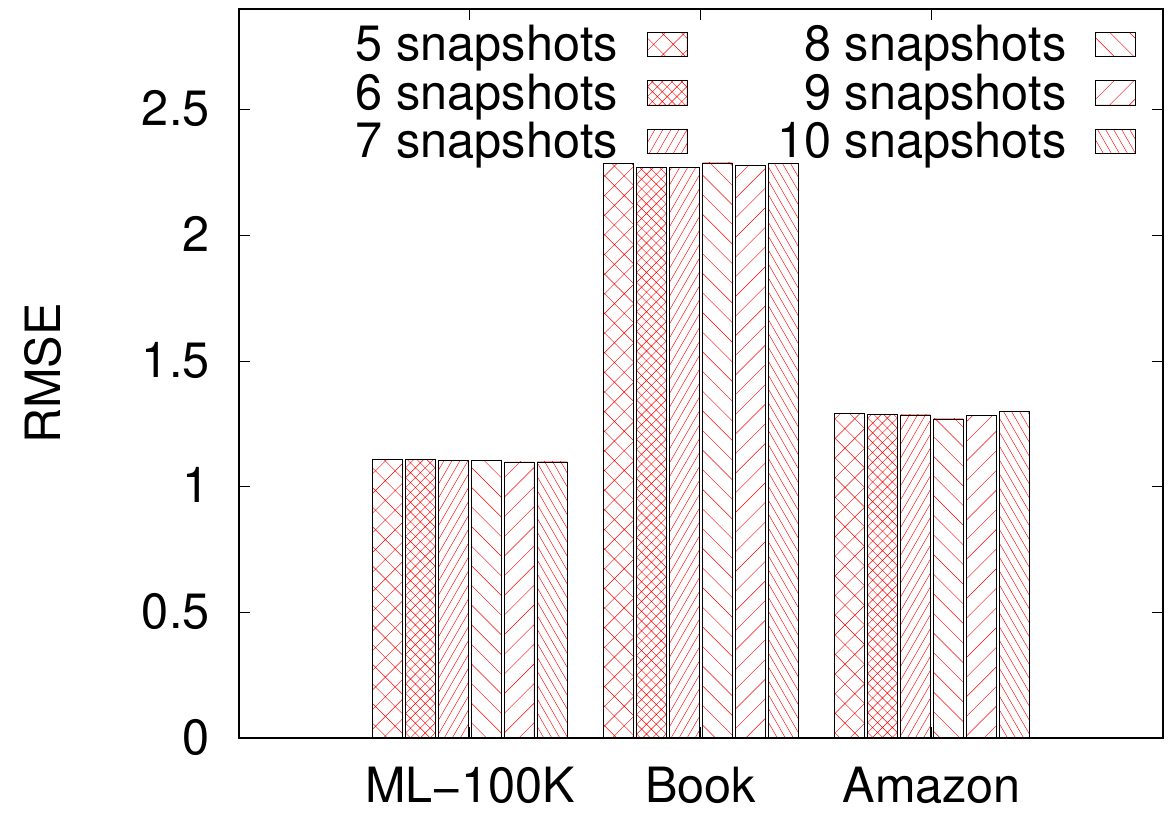}
		\caption{Sensitivity of NeuSE with different number of snapshot models over the RSVD method. The snapshot models are selected from the intermediate models trained every 10 epochs. We can see that selecting 7-9 snapshot models can achieve slightly better performance in this experiment.
			\label{fig:snapshot_num}}
	\end{minipage}
	\hfill
	\begin{minipage}[t]{0.32\linewidth}
		\centering
		\includegraphics[width=1\textwidth]{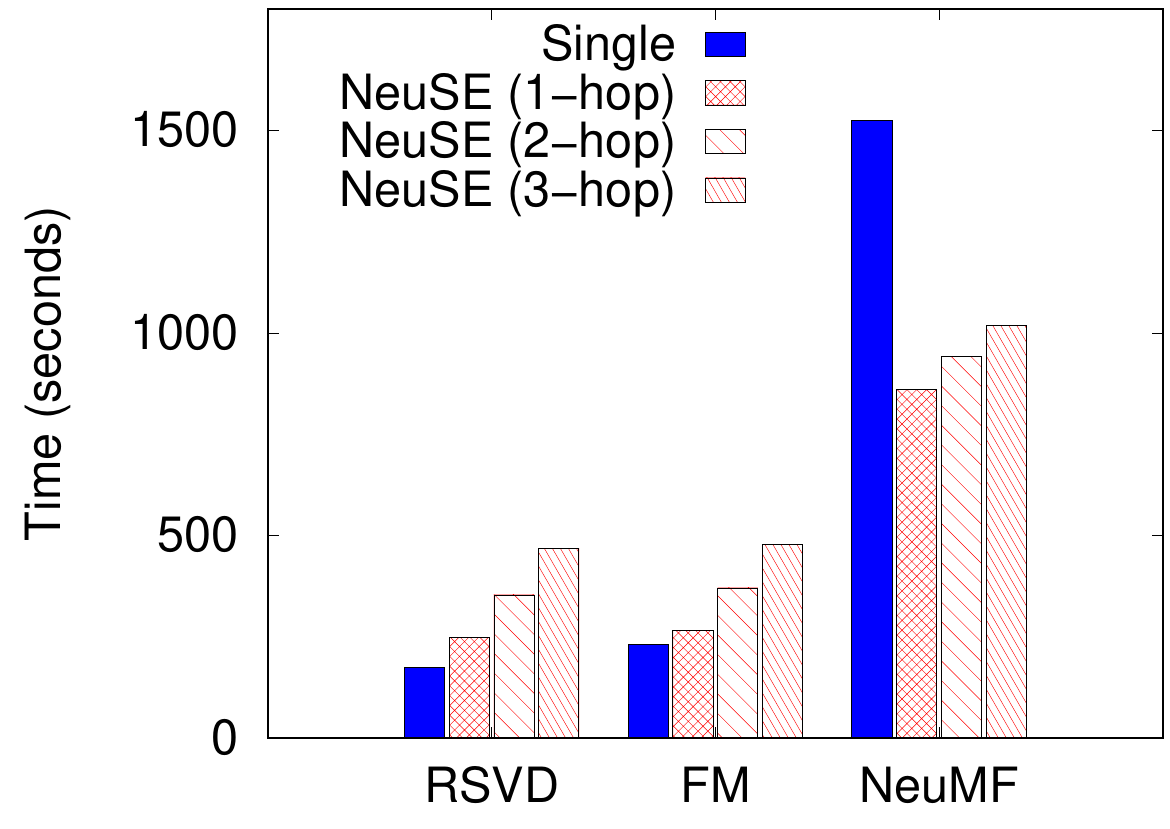}
		\caption{Efficiency comparison between training NeuSE models and training baseline CF models. We use ReLU as the activation function and 16 as embedding size in this experiment. The training time of NeuSE is comparable to training RSVD models and FM models and much lower than training NeuMF models.
			\label{fig:efficiency}}
	\end{minipage}
	\hfill
	\begin{minipage}[t]{0.32\linewidth}
		\centering
		\includegraphics[width=1\textwidth]{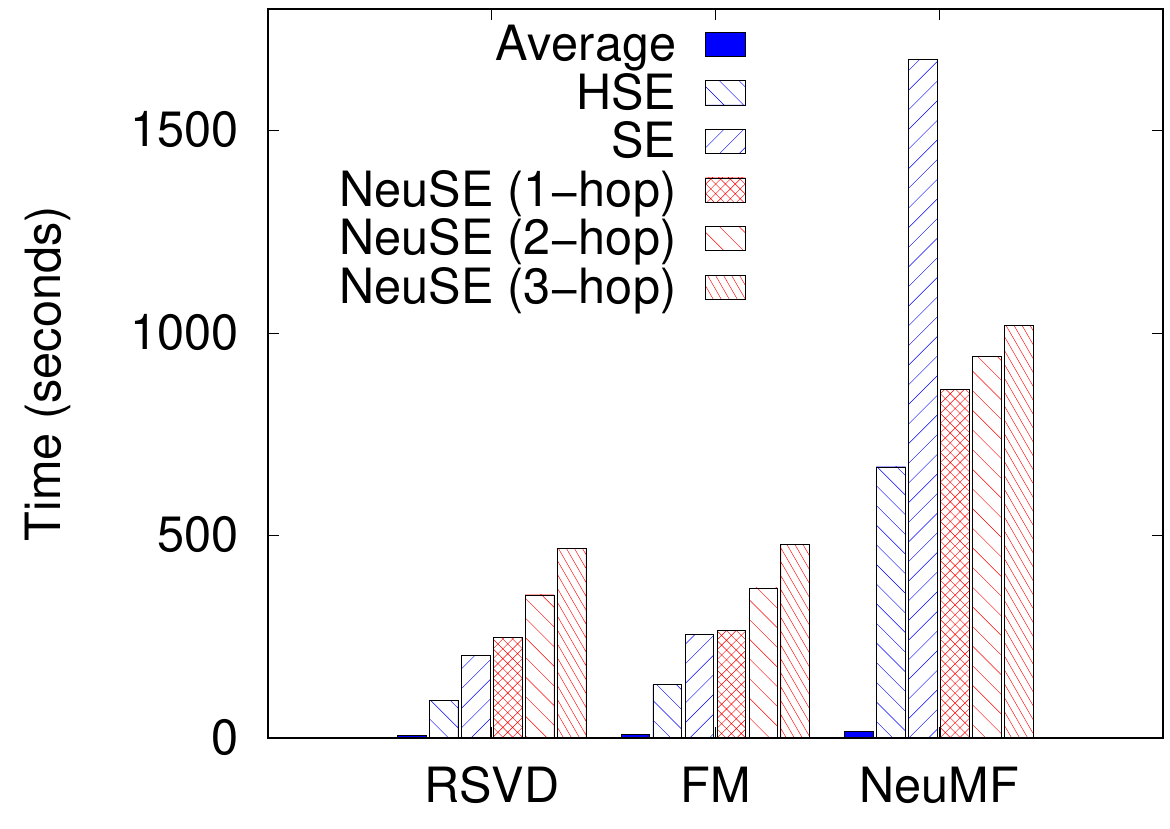}
		\caption{Efficiency comparison between NeuSE and other three snapshot ensemble methods. We use ReLU as the activation function and 16 as embedding size in this experiment. The efficiency of NeuSE is comparable to HSE and SE methods. The simple average method is much more efficient than other methods.
			\label{fig:efficiency_ensemble}}
	\end{minipage}
\end{figure*}

\subsubsection{Snapshot Selection}
Figure~\ref{fig:snapshot_num} shows the performance of NeuSE with different number of snapshot models over the RSVD method on three datesets. The snapshot models are selected from the intermediate models trained every 10 epochs. As we can see from the results, the performance variations are not very significant when we change the number of snapshot models from 5 to 10, which indicates that NeuSE can achieve decent performance even without optimal snapshot model selection. In addition, selecting 7-9 snapshot models can achieve slightly better performance than others in this experiment. This indicates that selecting too many snapshot models, \textit{e.g.,} 10, may hurt the accuracy, because some poor performance models may be selected. This conclusion is consistent with the observations from other ensemble learning works, in which researchers found that the ensemble performance may be hurt if the performance of the base models are too poor~\cite{Chen15,krogh1995neural, Zhou02}. However, thanks to the strong ensemble selection capability of NeuSE, the ensemble performance does not degrade much even we select 10 snapshot models.

\subsubsection{Efficiency Analysis}
Here, we first analyze the efficiency of applying NeuSE on three CF methods: RSVD, FM and NeuMF. As shown in Figure~\ref{fig:efficiency}, the efficiency of NeuSE is relevant to the baseline CF methods due to different learning trajectories of them but the computation time of NeuSE is comparable (RSVD, FM) or smaller (NeuMF) compared with the three CF methods. Therefore, we can conclude that NeuSE will not significantly increase the training time of CF methods compared with existing ensemble CF methods which train tens of base models~\cite{lee2013local,Chen15,min2018probabilistic}.

We also compare the additional computation overhead of NeuSE with other snapshot ensemble strategies, \textit{i.e.}, average, HSE and SE. As shown in Figure~\ref{fig:efficiency_ensemble}, the simple average method requires very little additional computation overhead due to no model learning steps and all the other snapshot ensemble methods have comparable computation overhead. Although NeuSE takes more additional computation overhead with increasing hops of memory networks, the computation time increases are linear. Therefore, we can conclude that NeuSE can achieve the best performance among all the snapshot ensemble methods without significantly increasing the computational overhead compared with HSE and SE.

\begin{table*}[t!]
	\centering
	\caption{Accuracy comparison between NeuSE and four snapshot ensemble methods (Single, Average~\cite{horizontal}, HSE~\cite{horizontal} and SE~\cite{snapshot}) on the Netflix Prize dataset. $N$ is set to 20 for HR@N and NDCG@N in the results. Average and HSE achieve similar performance with Single in all three algorithms. SE achieves better performance than Single, Average and HSE methods, but NeuSE outperforms all the compared methods in all cases. 
	\label{tab:netflix}}
	\begin{tabular}{c|c|ccc}
		\hline
		& Method & {RMSE} & {HR} & {NDCG}\\
		\hline
		\multirow{5}{*}{RSVD}
		& Single      & 1.1237 & - & - \\
		& Average     & 1.1238 & - & - \\
		& HSE		  & 1.1227 & - & - \\
		& SE          & 1.1205 & - & - \\
		& {\bf NeuSE} & {\textbf{1.1193}} & - & - \\
		\hline
		\multirow{5}{*}{FM}
		& Single      & 1.0737 & 0.6723 & 0.4830 \\
		& Average     & 1.0733 & 0.6718 & 0.4827 \\
		& HSE		  & 1.0741 & 0.6725 & 0.4833 \\
		& SE          & 1.0721 & 0.6798 & 0.4879 \\
		& {\bf NeuSE} & {\bf{1.0703}} & {\bf 0.6835} & {\bf 0.4918}\\
		\hline
		\multirow{4}{*}{NeuMF}
		& Single        & - & 0.8962 & 0.6254 \\
		& Average       & - & 0.8957 & 0.6246 \\
		& HSE			& - & 0.8983 & 0.6271 \\
		& SE            & - & 0.9021 & 0.6320 \\
		& {\bf NeuSE}& - & {\textbf{0.9073}} & { \textbf{0.6382}} \\
		\hline
	\end{tabular}
\end{table*}

\subsection{Performance on Large Dataset}
Next, we evaluate the performance of NeuSE on the Netflix Prize dataset to demonstrate its effectiveness on large datasets. The Netflix prize dataset consists of 100,480,507 ratings from 480,189 users given to 17,770 items. In this study, we compare both the accuracy and efficiency of NeuSE with the other baseline methods when applied on top of RSVD~\cite{paterek2007improving}, FM~\cite{FM} and NeuMF~\cite{NCF}. All the evaluation details are consistent with those described in Section~\ref{sec:exp_setup}.

Table~\ref{tab:netflix} compares the accuracy of NeuSE with other snapshot ensemble methods (Average, HSE and SE) on the Netflix Prize dataset on top of three popular CF algorithms (RSVD, FM (SGD) and NeuMF). As we can see from the results, NeuSE again outperforms all the other snapshot ensemble methods in all settings, which is consistent with the observations on relatively smaller datasets, e.g., MovieLens and Amazon. This confirms that NeuSE can boost the performance of collaborative filtering algorithms on both small datasets and large datasets.

\begin{figure}
	\centering
	\includegraphics[width=0.45\textwidth]{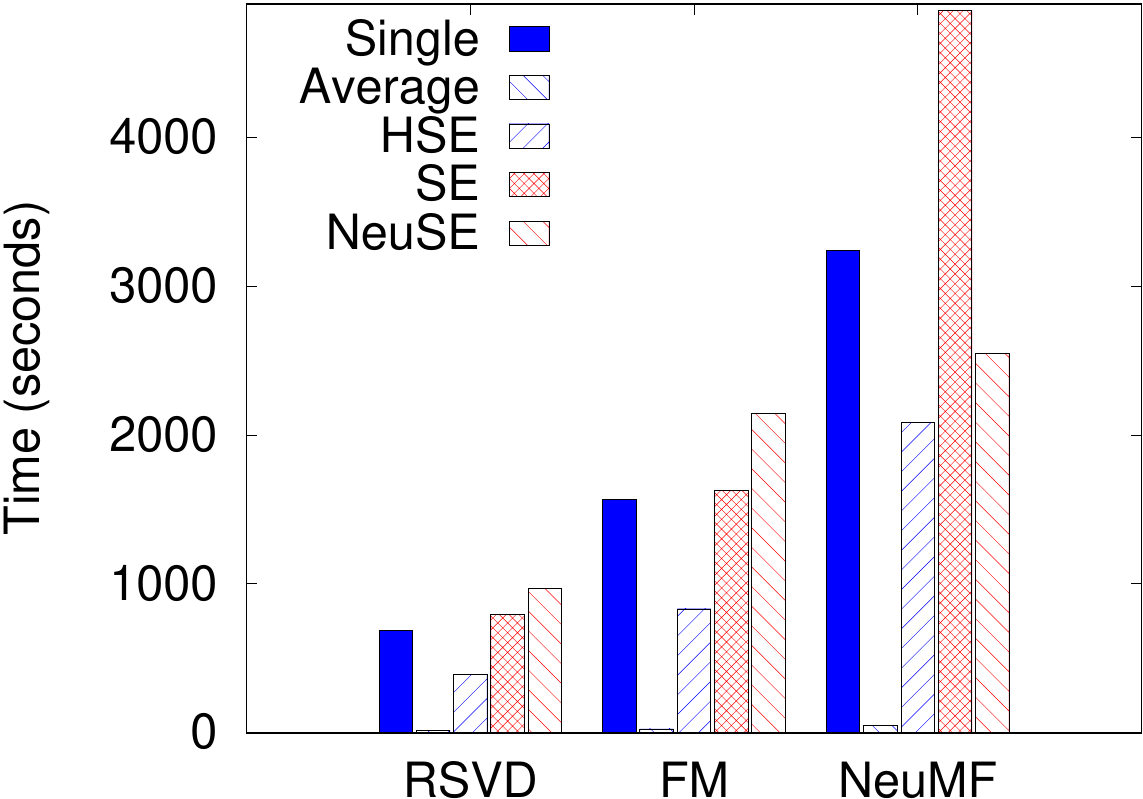}
	\caption{Efficiency comparison between NeuSE and other snapshot ensemble methods on the Netflix Prize dataset. }
	\label{fig:netflix_eff}
\end{figure}

Figure~\ref{fig:netflix_eff} compares the training time of NeuSE with other snapshot ensemble methods on the Netflix Prize dataset. As we can see from the results, the simple Average method is much more efficient than the other methods because no learning steps are involved. Similar to the comparison on the MovieLens datasets in Figure~\ref{fig:efficiency_ensemble}, the training overhead of NeuSE is on par with training a baseline CF model, which is also consistent with the observations on smaller datasets in Figure~\ref{fig:efficiency_ensemble}. Therefore, we can conclude that the training time of NeuSE is comparable to popular CF algorithms (e.g., RSVD~\cite{paterek2007improving} and NeuMF~\cite{NCF}) on both small datasets and large datasets.

In summary, we can conclude from the experimental results that NeuSE achieves decent performance (accuracy and efficiency) on both small datasets and large datasets.

\section{Conclusion and Future Work}

This paper proposes NeuSE -- a neural snapshot ensemble method for collaborative filtering algorithms. NeuSE integrates the snapshot models during global model learning via multi-memory networks, which only introduces a small amount of extra computational overhead but can extensively and significantly improve the performance of almost all kinds of existing collaborative filtering algorithms. One possible extension of this work is to apply the idea of NeuSE to other domains beyond collaborative filtering, e.g., non convex neural networks~\cite{choromanska2015loss}.

\section*{Acknowledgement}
This work is supported by National Natural Science Foundation of China (Project Nos. 61977013, 62090025) and Sichuan Science and Technology Program (Project No. 2019YJ0164).

\bibliographystyle{ACM-Reference-Format}
\bibliography{ref}

\end{document}